\documentclass[journal]{IEEEtran}

\usepackage{makecell}
\usepackage{times}
\usepackage{epsfig}
\usepackage{amssymb}
\usepackage{subfigure}
\usepackage{bbding}
\usepackage{xcolor}
\usepackage{threeparttable}
\usepackage{booktabs}
\usepackage{algorithm}
\usepackage{algorithmic}
\usepackage{url}

\newcommand{\eg}{\textit{e}.\textit{g}.}
\newcommand{\ie}{\textit{i}.\textit{e}.}
\newcommand{\et}{\textit{e}\textit{t} \textit{a}\textit{l}.}

\usepackage{graphicx, color, multirow, amsmath}

\RequirePackage{latexsym,amsmath,amssymb}

\ifCLASSINFOpdf
\else
\fi

\begin{document}

% \title{Subjective and Objective Quality Caption of Omnidirectional Images}
\title{Omnidirectional Image Quality Captioning: \\ A Large-scale Database and A New Model}
\author{
Jiebin Yan,
Ziwen Tan,
Yuming Fang,~\IEEEmembership{Senior~Member,~IEEE},
Junjie Chen,
Wenhui Jiang,\\
and Zhou Wang,~\IEEEmembership{Fellow,~IEEE}

% \thanks{This work was supported in part by the National Key Research and Development Program of China under Grant 2023YFE0210700, in part by the National Natural Science Foundation of China under Grants U24A20220, 62441203, 62461028, 62311530101, 62132006, and 62402201, in part by the Natural Science Foundation of Jiangxi Province of China under Grants 20243BCE51139, 20232BAB202001, and 20242BAB21006, in part by the project funded by China Postdoctoral Science Foundation under Grant 2024T170364, and the Project of the Education Department of Jiangxi Province of China under Grant GJJ2200522. (Corresponding author: Yuming Fang)}

\thanks{Jiebin Yan, Ziwen Tan, Yuming Fang, Junjie Chen and Wenhui Jiang are with the School of Computing and Artificial Intelligence, Jiangxi University of Finance and Economics, Nanchang 330032, Jiangxi, China. (e-mail: yanjiebin@jxufe.edu.cn, ziwentan@foxmai.com, fa0001ng@e.ntu.edu.sg).}

\thanks{Zhou Wang is with the Department of Electrical and Computer Engineering, University of Waterloo, Waterloo, ON, Canada (e-mail: zhou.wang@uwaterloo.ca).}
}

% \thanks{Copyright (c) 2022 IEEE. Personal use of this material is permitted. However, permission to use this material for any other purposes must be obtained by the IEEE by sending an email to pubs-permissions@ieee.org.}

% \markboth{Submitted to IEEE Transactions on Image Processing}%
% {Shell \MakeLowercase{\textit{et al.}}: Bare Demo of IEEEtran.cls for Journals}

%\markboth{IEEE Journal, 2019}
%{Shell \MakeLowercase{\textit{et al.}}: Bare Demo of IEEEtran.cls for Journals}
% after the title page when using the two sides option.
\maketitle

\begin{abstract}
The fast growing application of omnidirectional images calls for effective approaches for omnidirectional image quality assessment (OIQA). Existing OIQA methods have been developed and tested on \textit{homogeneously} distorted omnidirectional images, but it is hard to transfer their success directly to the \textit{heterogeneously} distorted omnidirectional images. In this paper, we conduct the largest study so far on OIQA, where we establish a large-scale database called OIQ-10K containing 10,000 omnidirectional images with both homogeneous and heterogeneous distortions. A comprehensive psychophysical study is elaborated to collect human opinions for each omnidirectional image, together with the spatial distributions (within local regions or globally) of distortions, and the head and eye movements of the subjects. Furthermore, we propose a novel multitask-derived adaptive feature-tailoring OIQA model named IQCaption360, which is capable of generating a quality caption for an omnidirectional image in a manner of textual template. Extensive experiments demonstrate the effectiveness of IQCaption360, which outperforms state-of-the-art methods by a significant margin on the proposed OIQ-10K database. The OIQ-10K database and the related source codes are available at \url{https://github.com/WenJuing/IQCaption360}.
\end{abstract}
% This preserves the distinction between vectors and scalars. However, if the journal you are submitting to favors bold math in the abstract,
% then you can use LaTeX's standard command \boldmath at the very start of the abstract to achieve this. Many IEEE journals frown on math

% Note that keywords are not normally used for preview papers.

% For preview papers, this IEEE trans. command inserts a page break and
\IEEEpeerreviewmaketitle

\section{Introduction}
\label{sec:intr}

\IEEEPARstart{V}{irtual} reality (VR) is characterized by providing a simulated environment where end users can freely enjoy an immersive experience. The omnidirectional image, also called 360$^{\circ}$ image and panorama image, acts as an important medium of visual representation of VR and may degrade in the procedure of acquisition, transmission, processing, storage~\cite{wang2017begin}. The degradation of omnidirectional images not only greatly impairs the quality of experience (QoE) of end users, but also affects the usability of omnidirectional images~\cite{vidalmata2020bridging}. Estimating the visual quality of omnidirectional images accurately is of great importance for algorithm development and system optimization~\cite{fang2021superpixel,ding2021comparison}. Omnidirectional image quality assessment (OIQA) consists of subjective and objective quality assessment, where the former type of study conducts psychophysical experiments to investigate the influence of various factors and establishes benchmarks for comparing different objective models. Generally, objective OIQA models can be roughly categorized into full-reference OIQA (FR-OIQA), reduced-reference OIQA (RR-OIQA), and no-reference/blind OIQA (NR-OIQA) according to the availability of reference images. The former two types of methods estimate the quality of omnidirectional images using full and partial reference information, respectively, while the NR-OIQA models can be deployed without access to reference information.

\begin{figure}[t]
\centering
\subfigure[\textbf{QV:} 2.72; \textbf{QC:} A \emph{good}-quality omnidirectional image with \emph{no perceptibly distorted region}. It \emph{should be saved}.]{
\includegraphics[width=0.46\linewidth]{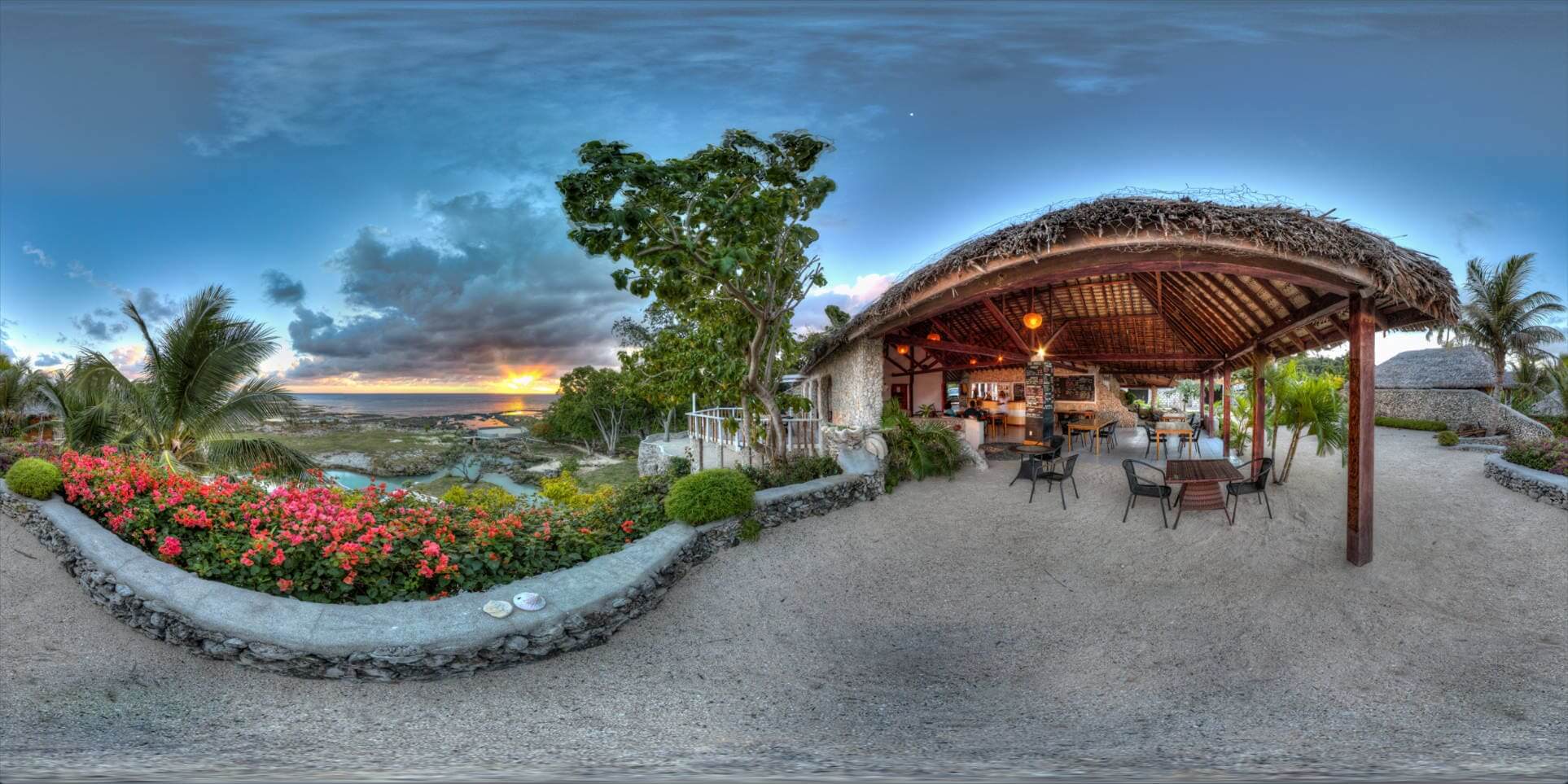}
}
\subfigure[\textbf{QV:} 2.17; \textbf{QC:} A \emph{fair}-quality omnidirectional image with \emph{one distorted region}. It \emph{is recommended to be saved}.]{
\includegraphics[width=0.46\linewidth]{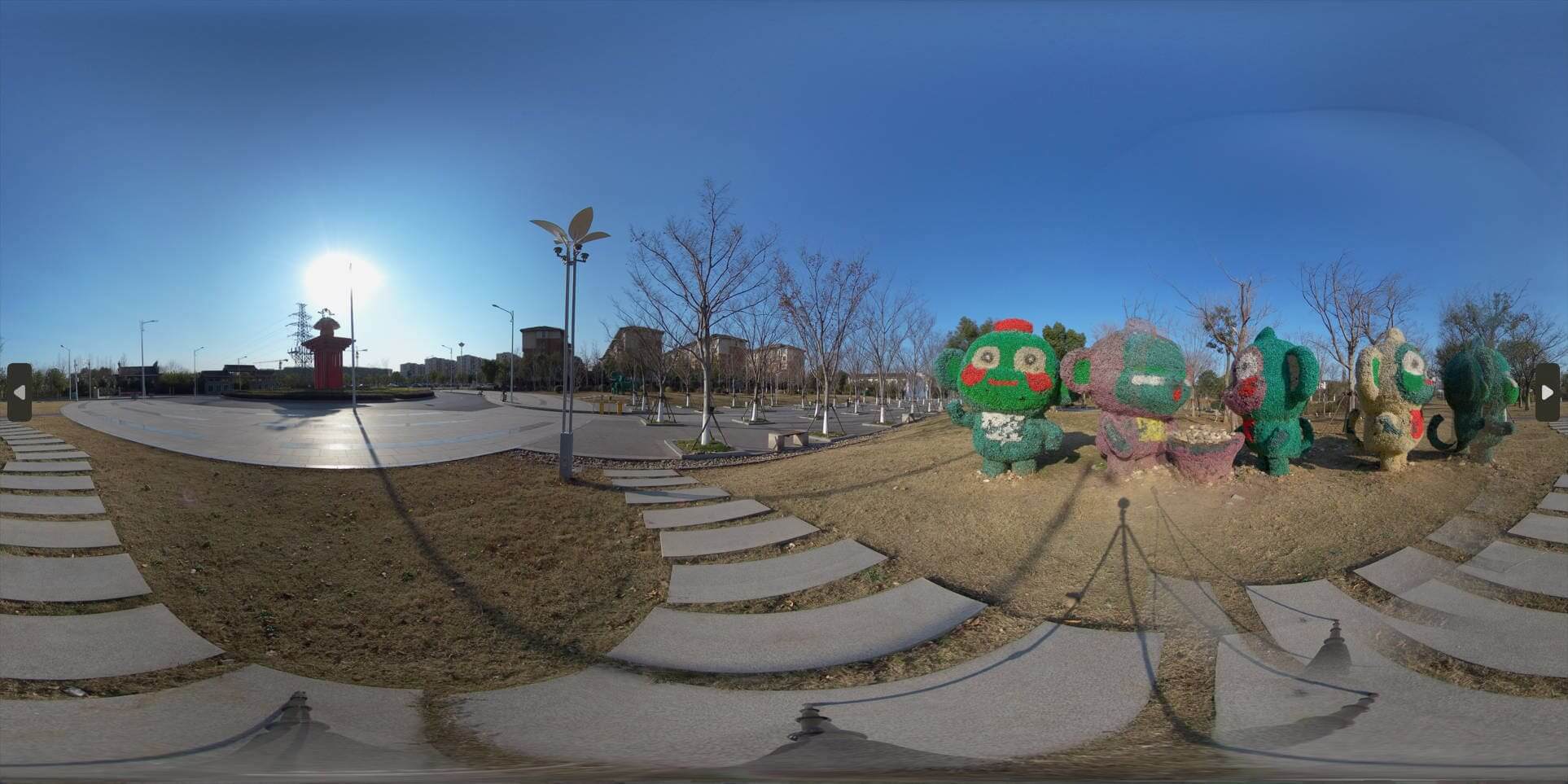}
}
\subfigure[\textbf{QV:} 1.80; \textbf{QC:} A \emph{fair}-quality omnidirectional image with \emph{two distorted regions}. It \emph{is recommended to be discarded}.]{
\includegraphics[width=0.46\linewidth]{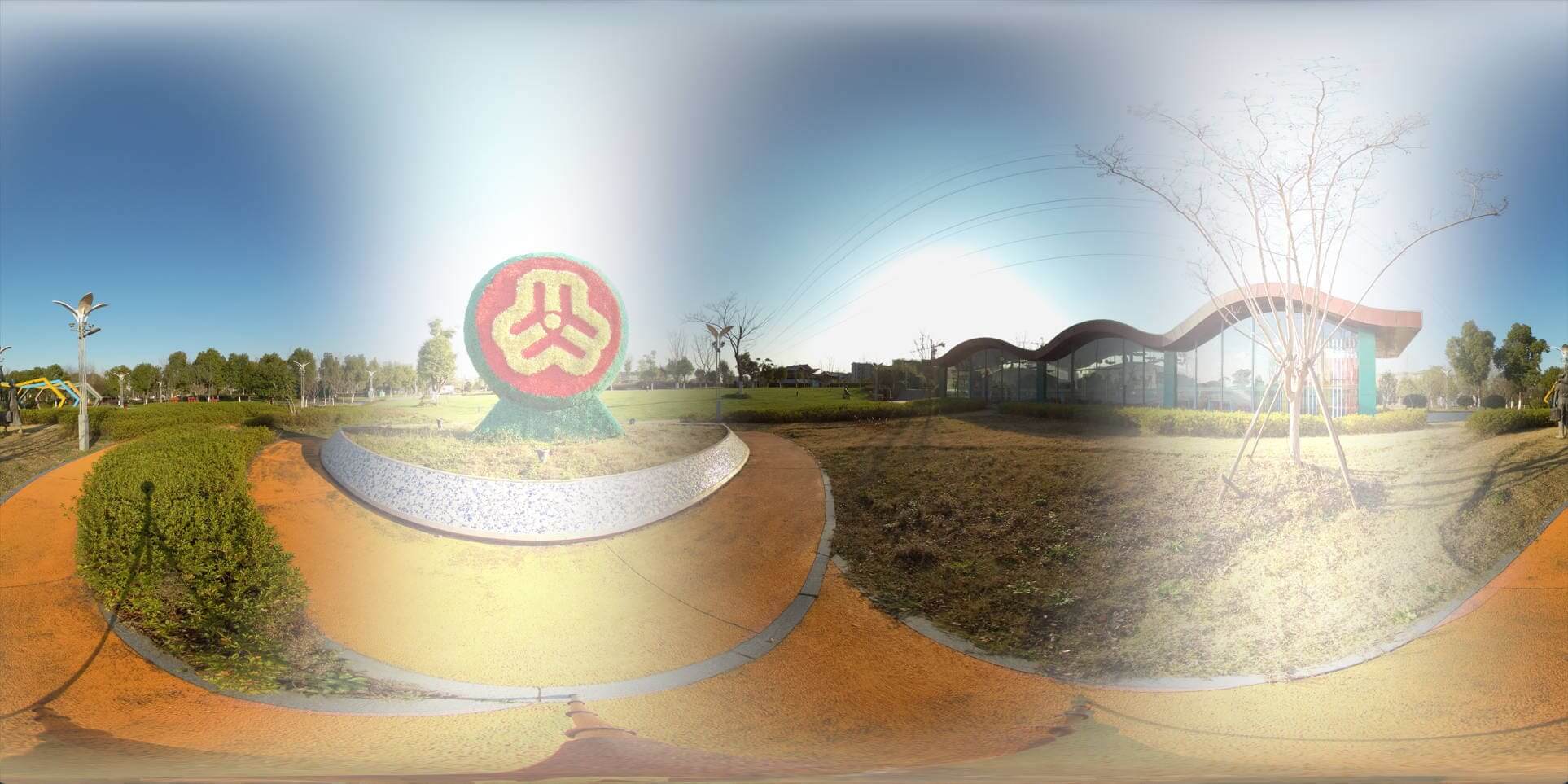}
}
\subfigure[\textbf{QV:} 1.00; \textbf{QC:} A \emph{poor}-quality omnidirectional image with \emph{global distortion}. It \emph{should be discarded}.]{
\includegraphics[width=0.46\linewidth]{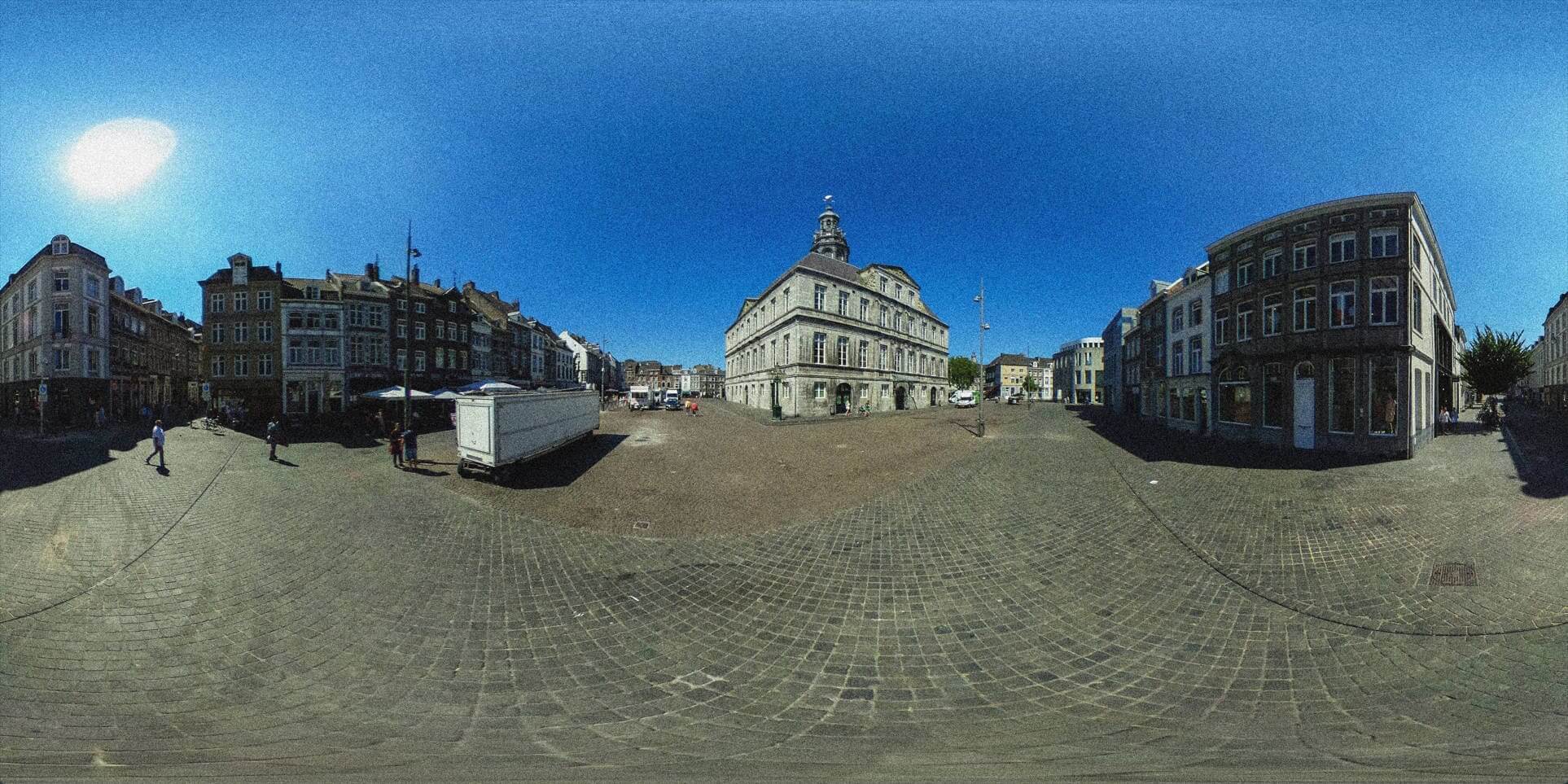}
}
\caption{Visual examples represented by quality value and quality caption.~\textbf{QV} represents \textbf{Q}uality~\textbf{V}alue, and~\textbf{QC} represents~\textbf{Q}uality~\textbf{C}aption.}
\label{fig:motivation}
\end{figure}

\begin{table*}[t]
\centering
\caption{Comparison of existing OIQA databases and the proposed OIQ-10K database.}
\begin{tabular}{lcccccp{4cm}}
\toprule
\textbf{Database} & \textbf{\# Year} & \textbf{\# Ref./Dist.} & \textbf{\# Number of images} & \textbf{\# Degradation type} & \textbf{\# Subjects} & \textbf{\makecell[c]{Public availability}} \\
\midrule
% 1 & Salient360 \cite{sun2018large} & 2017 & 98 & 20 & Uniform & MOS \\
CVIQ \cite{sun2018large} & 2018 & 16/528 & 544 & Homogeneous & 20 &  \makecell[c]{https://github.com/\\sunwei925/CVIQDatabase} \\
\midrule
OIQA \cite{duan2018perceptual} & 2018 & 16/320 & 336 & Homogeneous & 20 & \makecell[c]{Upon request} \\
\midrule
LIVE 3D VR IQA \cite{chen2019study} & 2019 & 15/450 & 465 & Homogeneous & 42 & \makecell[c]{http://live.ece.utexas.edu/\\research/VR3D/index.html} \\
\midrule
MVAQD \cite{jiang2021cubemap} & 2021 & 15/300 & 315 & Homogeneous & 26 & \makecell[c]{Upon request} \\
\midrule
NBU-HOID \cite{cao2021quality} & 2021 & 16/320 & 336 & Homogeneous & 31 & \makecell[c]{https://github.com/\\caoliuyan/NBU-HOID} \\
\midrule
NBU-SOID \cite{qi2020viewport} & 2021 & 12/396 & 408 & Homogeneous & 30 & \makecell[c]{https://github.com/\\qyb123/NBU-SOID/} \\
\midrule
% 6 & AOI \cite{xu2021saliency} & 2021 & 600 & 30 & Uniform & HMData \\
IQA-ODI \cite{yang2021spatial} & 2021 & 120/960 & 1,080 & Homogeneous & 200 & \makecell[c]{https://github.com/\\yanglixiaoshen/SAP-Net} \\
\midrule
JUFE \cite{fang2022perceptual} & 2021 & 258/1,032 & 1,290 & Heterogeneous & 120 & \makecell[c]{https://github.com/\\LXLHXL123/JUFE-VRIQA} \\
\midrule
OIQ-10K & 2024 & 2,500/7,500 & \textbf{10,000} & \textbf{\makecell[c]{Homogeneous and \\Heterogeneous}} & 160 & \makecell[c]{https://github.com/\\WenJuing/IQCaption360} \\
\bottomrule
\end{tabular}

\label{tab:databases_summary}
\end{table*}

Some early methods~\cite{yu2015framework,sun2017weighted,zakharchenko2016quality} are mainly modified from two classic IQA models,~\ie, PSNR and structure similarity (SSIM)~\cite{wang2004image}. However, their performance is suboptimal mainly due to the complex immersive interaction between users and visual content. Recently, many deep neural network (DNN) based methods~\cite{sun2019mc360iqa,kim2019deep,xu2020blind,fu2022adaptive,wu2023assessor360} have been proposed and achieved promising results on existing OIQA databases by virtue of end-to-end learning. Nevertheless, there are still two drawbacks in OIQA that need to be addressed: $\mathbf{\romannumeral1)}$~\textbf{Limited diversity in image content and distortion variation}. All existing OIQA databases contain a limited number of image samples compared to 2D-IQA databases~\cite{fang2020perceptual,hosu2020koniq}, which constrains the generalization ability of these learning-based methods. Besides, the omnidirectional images in most existing OIQA databases except the JUFE database~\cite{fang2022perceptual} suffer only from global distortion (\ie, homogeneous distortion). That is, all regions in each omnidirectional image suffer from the same type of distortion with the same level, which is inconsistent with the realistic scenarios since non-synchronization between different lenses of the VR camera or something wrong with one or some of the lenses can lead to locally distributed distortion (\ie, heterogeneous distortion)~\cite{fang2022perceptual,yan2022subjective}. $\mathbf{\romannumeral2)}$~\textbf{Single quality representation}. Different from other computer vision tasks, IQA is a relatively subjective task~\cite{zhang2021uncertainty},~\ie, it is not easy to assign a certain quality score to an image. Zhang~\et~\cite{zhang2021uncertainty} proposed a unified IQA framework, which predicts the mean quality value and quality variance (which can be regarded as the measurement of the \textit{uncertainty} about image quality) of the input image. In~\cite{yang2022fine}, Yang~\et~considered that humans prefer to give a semantic description rather than a quantitative quality value, and proposed a human cognition-oriented IQA model, which focuses on hierarchical semantics degradation. They designed three indexes to describe hierarchical semantics together, including local-detail semantics (edge), regional structure semantics (contours), and global-concept semantics (categories)~\cite{yang2022fine}. However, almost all existing OIQA studies are dedicated to designing an efficient mapping function from an omnidirectional image to a single quality score, ignoring the sophistication of human QoE assessment.

To address the above problems, we conduct a large-scale study on quality assessment of both homogeneously and heterogeneously distorted omnidirectional images and aim to build an objective model that can automatically generate a quality caption for a test omnidirectional image. Moreover, different from our previous study~\cite{fang2022perceptual} which investigates the OIQA problem from users' perspectives (\ie, viewing condition and viewing behavior), here we opt for a holistic perspective (\ie, global quality degradation and distortion distribution). To this end, we consider the overall quality of images rather than the conditional quality in \cite{fang2022perceptual}, and a broader diversity of distortions is considered in this work. To better show our idea, we give an intuitive example in Figure~\ref{fig:motivation}, where four omnidirectional images with different quality scores and their quality-aware descriptions are presented. Rather than providing a scalar value for representing the quality of omnidirectional images, our objective is to provide a more informative quality-aware description for omnidirectional images, and we refer to this problem as~\textbf{Omnidirectional Image Quality Captioning (OIQC)}.

Specifically, we first develop a new OIQA database named OIQ-10K which contains four distortion situations based on the distortion distribution, and a large-scale psychophysical experiment is conducted to collect human opinions on image quality and the spatial distribution of image distortions. Furthermore, we propose a novel NR-OIQA model called IQCaption360 to deal with the OIQC problem. Considering that the vision transformer (ViT)~\cite{dosovitskiy2020image} exhibits effective global feature extraction power through introducing the attention mechanism, therefore we adopt a transformer-based network to extract multi-scale features. We formulate the OIQC problem as a multi-task framework, where a distortion situation prediction network is designed to measure the global degradation of omnidirectional images and a quality score prediction network to estimate the quality of omnidirectional images. Since different tasks rely on specific features~\cite{liu2020dynamic}, we design a multitask-derived feature selection mechanism to adaptively allocate task-specific features to the aforementioned two networks.

In summary, our contributions are three folds:
\begin{itemize}
\item We construct so far the largest OIQA database called OIQ-10K, which contains 10,000 omnidirectional images with homogeneous and heterogeneous distortion. Additionally, the OIQ-10K database contains rich content and complex degradation situations. 

\item A large-scale psychophysical study is carefully elaborated to collect human opinions for omnidirectional images, together with head and eye movement. Specifically, we employ a coarse-to-fine strategy to expedite subjective annotation, and detailedly analyze the influence of distortion situation and viewing condition on perceptual quality on the consistency of subjects' QoE.

\item We propose a novel multitask-derived adaptive feature-tailoring model named IQCaption360, which is able to capture distortions with different ranges and produce quality captions for omnidirectional images in a manner of textual template. 

\end{itemize}

\section{Related Work}
\label{sec:rw}

In this section, we first briefly introduce existing OIQA databases. Then, we describe state-of-the-art OIQA models and IQC models.

\subsection{Subjective OIQA Databases}
\label{subsec:database}

The construction of IQA databases needs much more effort than establishing benchmarks for other computer vision tasks, \eg, object detection, semantic segmentation, and \emph{etc}, since IQA is a relatively uncertain problem and subjective quality assessment requires more labor to get stable ratings~\cite{yan2022subjective,zhang2021uncertainty}. Generally, IQA databases act as the platform for investigating the influence of various factors on QoE of users and comparing different IQA models. Compared with the 2D-IQA databases, the weaknesses of existing OIQA databases include at least a limited number of image samples and simplistic content. The CVIQ database~\cite{sun2018large} is composed of 528 images with compression distortion, which are generated from 16 reference images, while the images in the OIQA database~\cite{duan2018perceptual} are distorted by four types of synthesized distortions. Similarly, the LIVE VR IQA database~\cite{chen2019study} includes 450 distorted images generated from 15 reference images, where the images are distorted by six types of synthesized distortions; the MVAQD database~\cite{jiang2021cubemap} consists of 15 source images and 300 distorted images with five types of synthesized distortions; the NBU-HOID database~\cite{cao2021quality} investigates the influence of compression and tone mapping on the quality of high dynamic range omnidirectional images; the NBU-SOID database~\cite{qi2020viewport} focuses on quality assessment of stereoscopic omnidirectional images, each of which is composed of a left omnidirectional image and a right omnidirectional image; the IQA-ODI database~\cite{yang2021spatial} explores the impairments of JPEG compression and map projection on omnidirectional images. 

To sum up, all the databases mentioned above are dedicated to the issue of homogeneous degradation, \ie, every region suffers from the same type of distortion with the same level regardless of its position. These databases undeniably promote the development of OIQA, however, they ignore the heterogeneous degradation problem, which is also very important in real applications~\cite{yan2022subjective}. Fang~\et~\cite{fang2022perceptual} introduced the concept of heterogeneous distortion and built an OIQA database named JUFE (which is extended on the small database~\cite{sui2021perceptual}), where they discussed the relationship between viewing condition and viewing behavior. Nevertheless, the few number of omnidirectional images in the JUFE database makes it difficult to satisfy the data-hungry demand of deep learning models. Thus, to further facilitate the development of OIQA, we propose a new large-scale database named OIQ-10K, which consists of 10,000 omnidirectional images. Besides, to approximate real applications, the proposed OIQ-10K database contains not only homogeneous distortion, but also heterogeneous distortion. The comparison of the proposed OIQ-10K database and the existing databases is listed in Table~\ref{tab:databases_summary}.

\subsection{Objective OIQA Models}
\label{subsec:oiqa_method}

A straightforward method to estimate the quality of omnidirectional images is to apply or slightly modify 2D-IQA models according to the sphere characteristic of images, such as PSNR-based models \cite{yu2015framework,sun2017weighted,zakharchenko2016quality} and SSIM-based models \cite{zhou2018weighted}. This type of model is far from being consistent with human ratings due to their limited quality-aware representation ability and the overlook of human behaviors~\cite{fang2022perceptual, wu2023assessor360}. Recently, many deep-learning-based methods have been proposed and shown promising performance. Some of them accept the omnidirectional image in the format of equirectangular projection (ERP) or cubemap projection (CMP) \cite{kim2019deep,chai2021monocular} directly. Considering the limited field of view, several researchers introduced the concept of viewport to design OIQA models. Sun~\et~\cite{sun2019mc360iqa} extracted six viewports from each omnidirectional image to cover the full visual content and enlarge the training samples by repeating this process with varying starting angles. Xu~\et~\cite{xu2020blind} and Fu~\et~\cite{fu2022adaptive} adopted the graph neural network to aggregate viewport quality. Other than using a viewport sequence to represent an omnidirectional image, some studies have also considered users' viewing behavior. Yang~\et~\cite{yang2022tvformer} proposed a trajectory-guided OIQA model, which is jointly optimized by OIQA and head trajectory prediction. Wu~\et~\cite{wu2023assessor360} designed a recursive probability sampling scheme to simulate users' browsing process and generate multiple pseudo viewport sequences from a given starting point.

Overall, these OIQA models achieve great results in capturing the quality of homogeneously distorted omnidirectional images; however, they overlook the exploration of heterogeneously distorted omnidirectional images that exist in the real world. In addition, a single quality representation, \emph{i.e.}, relying on a value to characterize image quality, could limit its ability in practical application.
\subsection{IQC Models}
\label{subsec:iqc_method}

Image captioning, which bridges computer vision and natural language processing using artificial intelligence techniques, aims to automatically generate a description about the \textit{content} and \textit{semantic} of an image~\cite{vinyals2015show}, \eg, \textit{a person is sitting on a chair next to a very small house}, rather than a global-wise single category or a pixel-wise semantic classification map~\cite{yan2021exposing}. Likewise, the objective of IQC is to describe the \textit{visual quality} using one sentence, \eg, \textit{local details are observably distorted, regional contours are slightly distorted, and global concepts are basically undistorted}~\cite{yang2021image}, which has richer information about visual quality than a single quantitative value. Following this idea, Yang~\et~\cite{yang2021image} presented an initial study on IQC by considering it as a needs-oriented task and designed an attentive and recurrent semantic attractor network to address this problem. Latter, Yang~\et~\cite{yang2022fine}~extended their previous work~\cite{yang2021image} in a bi-directional manner. Note that IQC in both of these two works is formulated as a multiclass classification problem, and the quality description is generated by a feat of the pre-defined textual template. Similarly, Zhang~\et~\cite{zhang2023blind} proposed to use the pre-defined textual template to construct the vision-language correspondence and optimize the joint probability over multiple tasks. Except for using hand-crafted templates to promote the learning of quality-aware visual features, some researchers have also attempted to use the \textit{users' comments} to assist in model training~\cite{nie2023bmi,zhong2023aesthetically}.

\begin{figure}[t]
\centering
\subfigure[CVIQD]{
\includegraphics[width=0.29\linewidth]{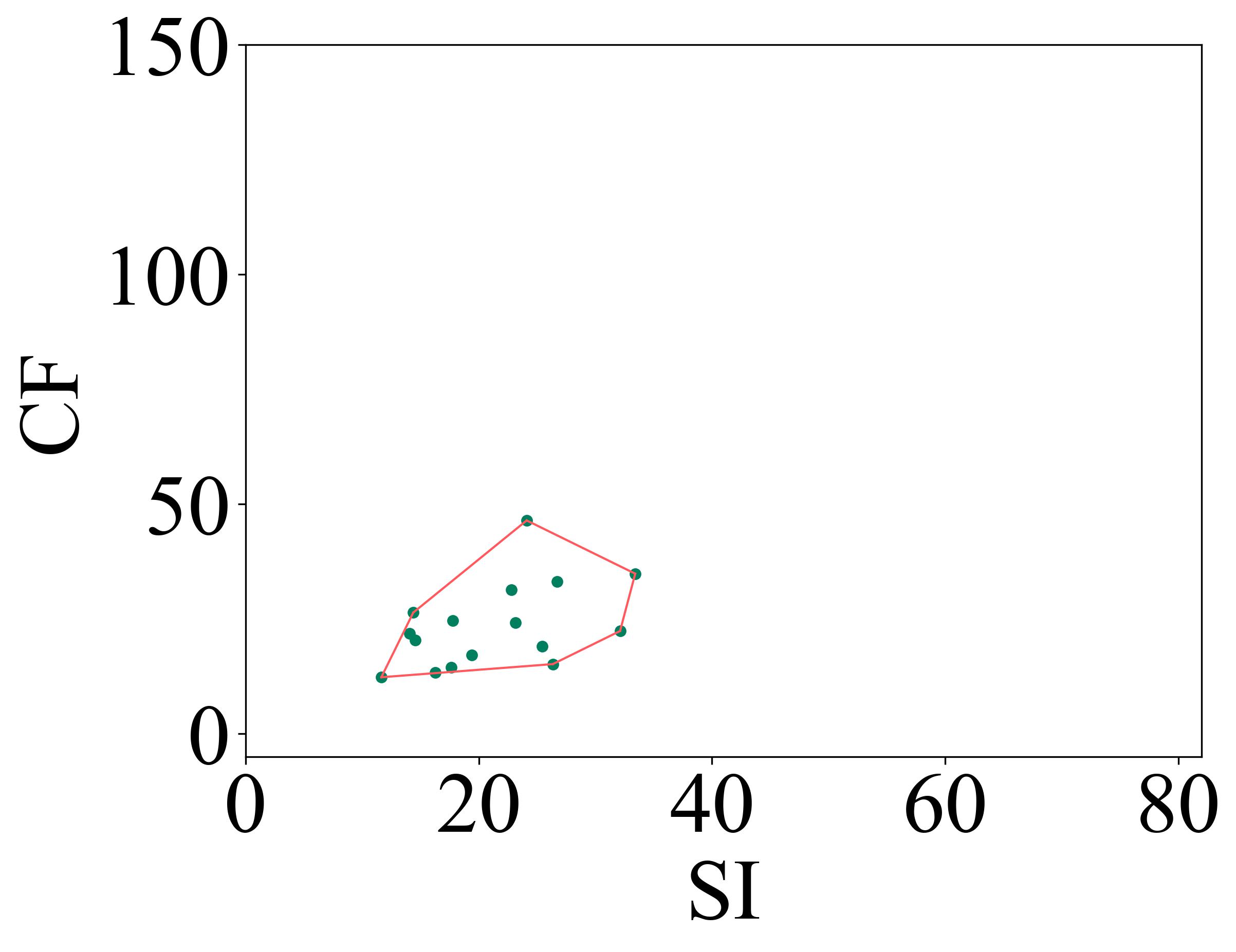}
}
\subfigure[OIQA]{
\includegraphics[width=0.29\linewidth]{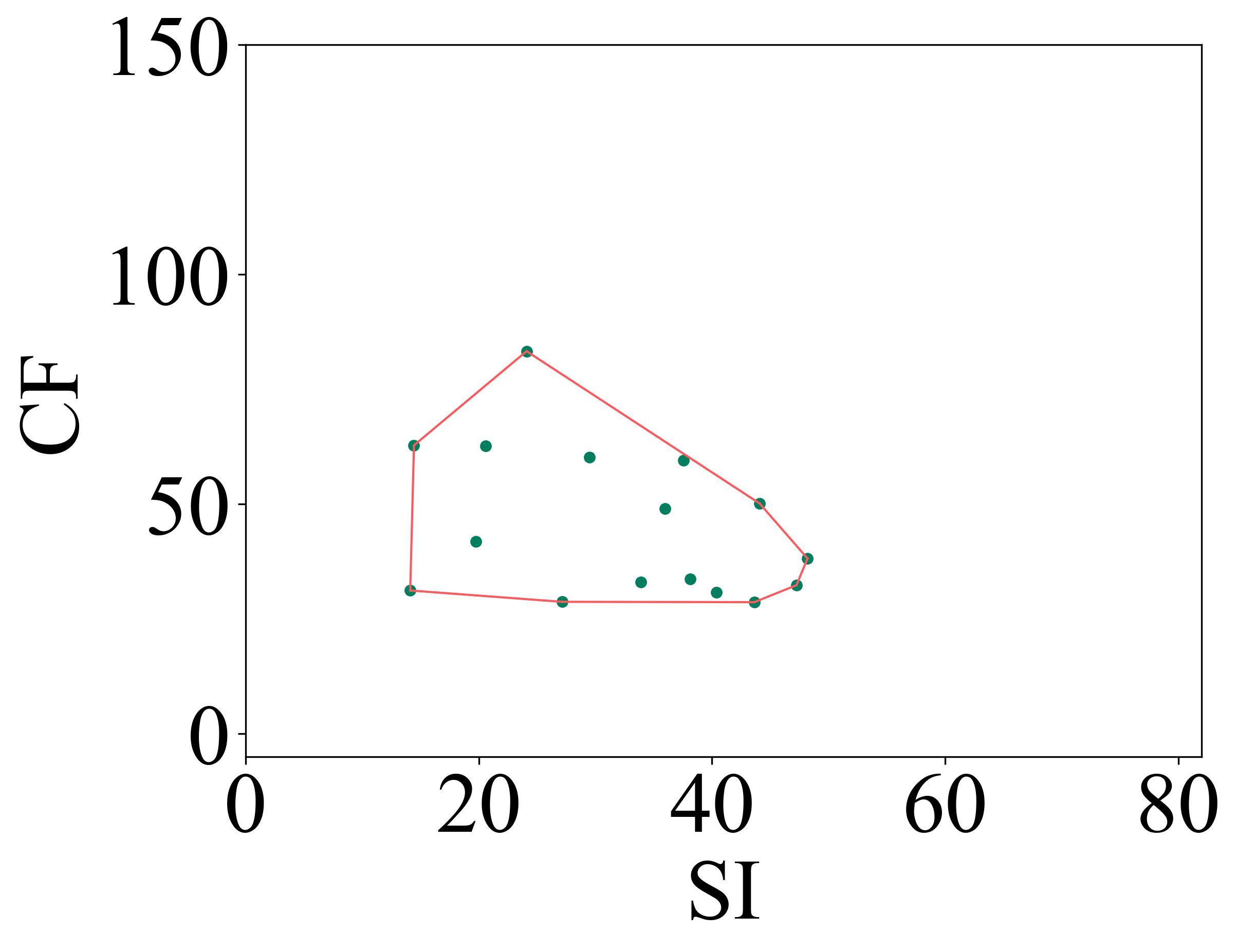}
}
\subfigure[LIVE 3D VR IQA]{
\includegraphics[width=0.29\linewidth]{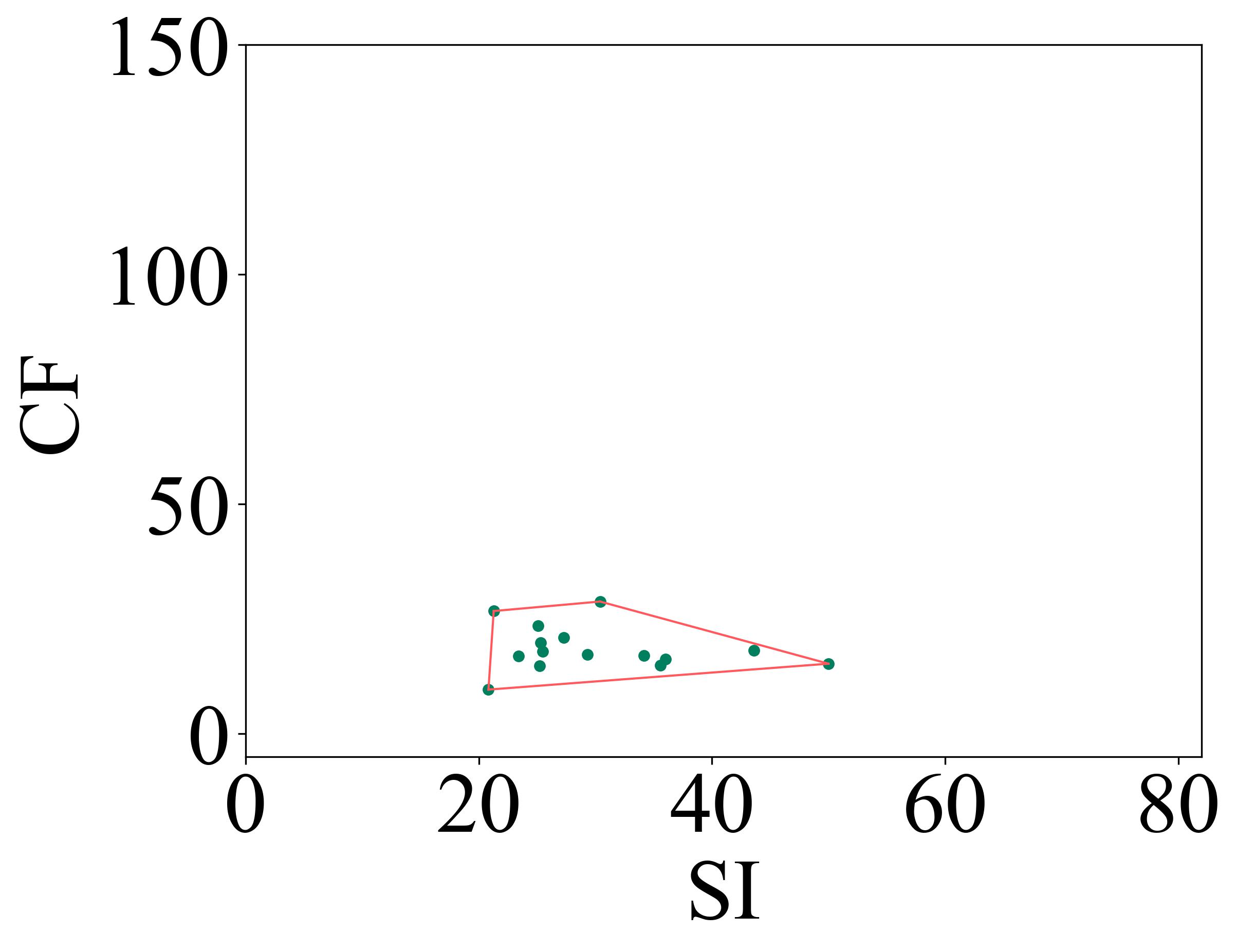}
}
\subfigure[MVAQD]{
\includegraphics[width=0.29\linewidth]{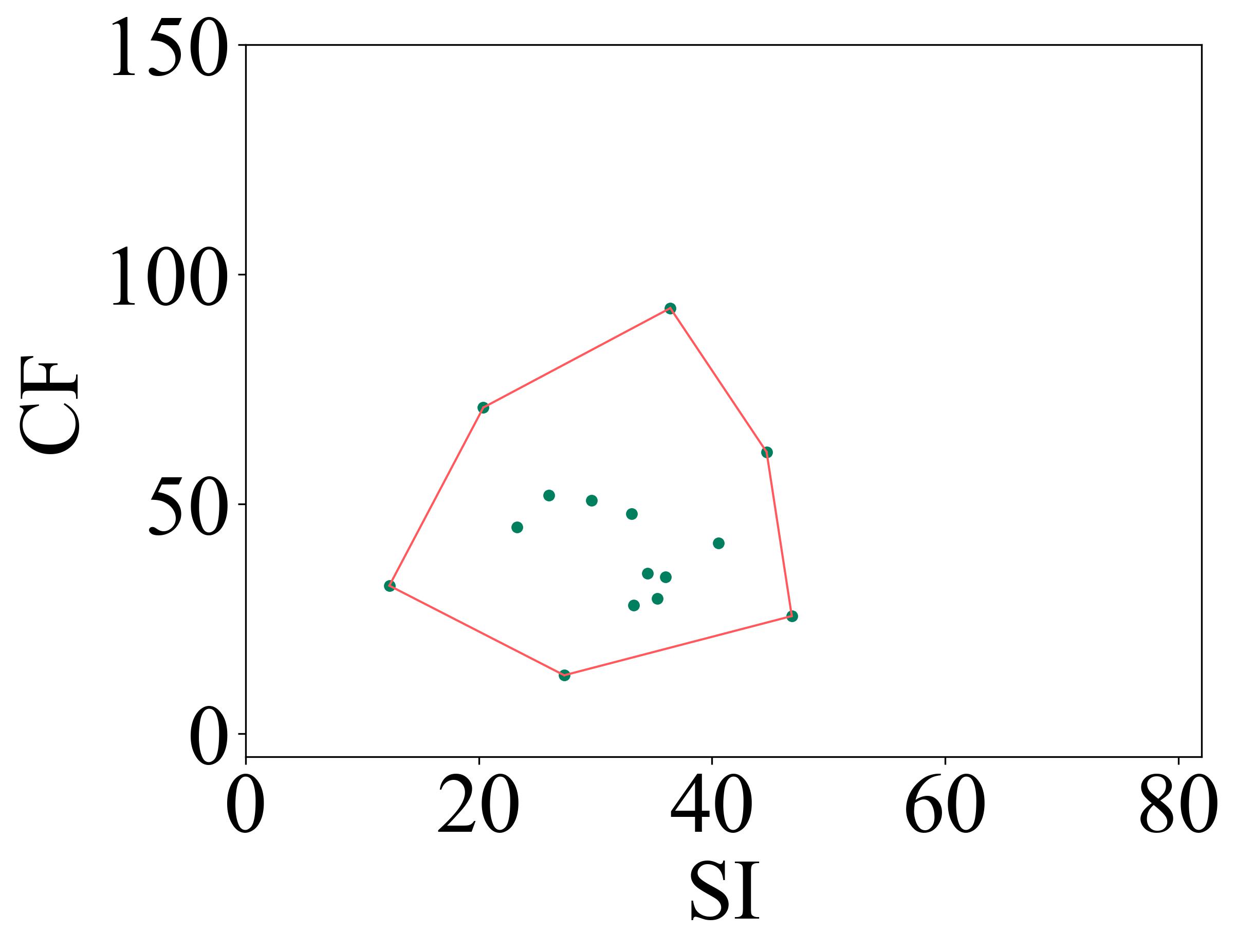}
}
\subfigure[NBU-SOID]{
\includegraphics[width=0.29\linewidth]{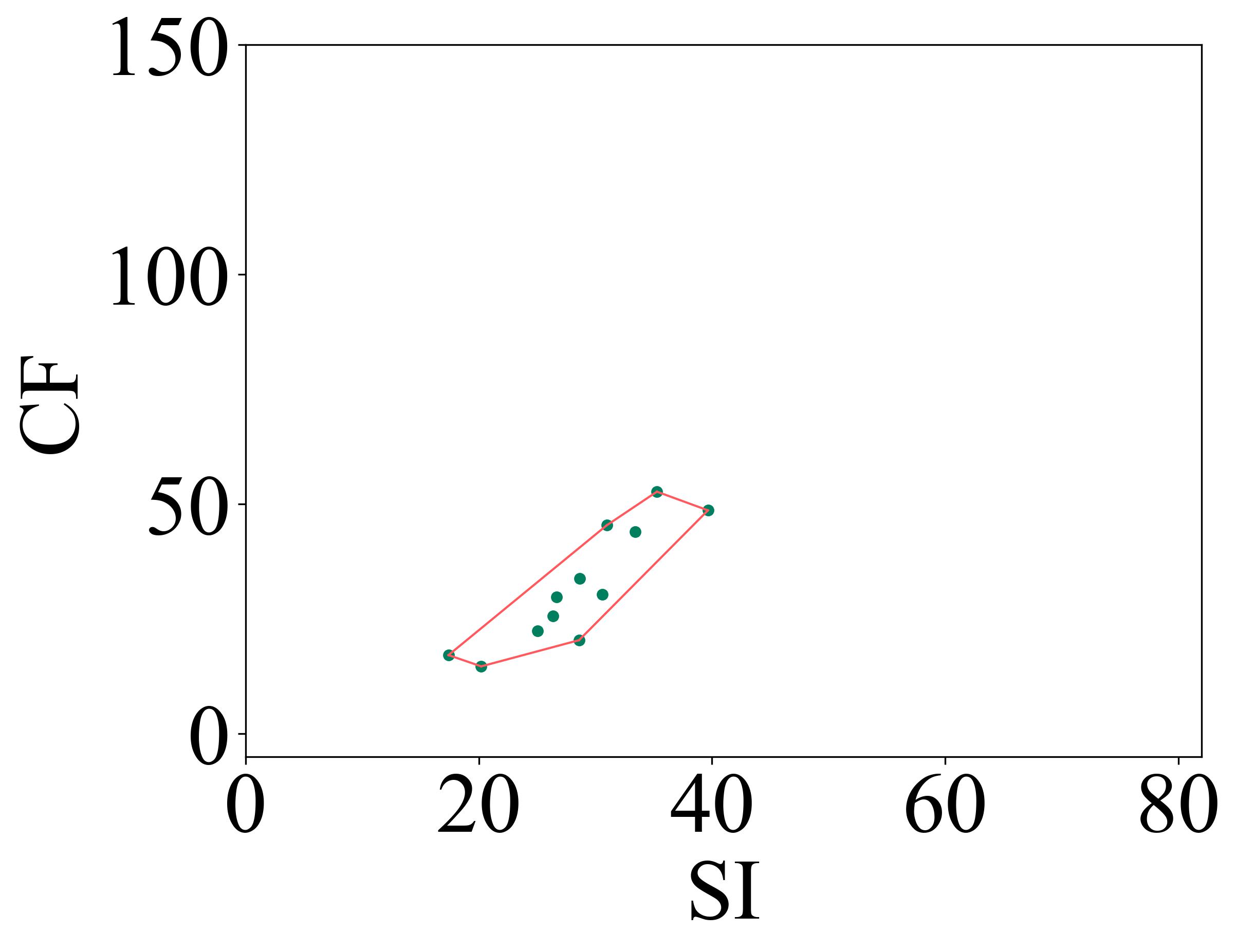}
}
\subfigure[IQA-ODI]{
\includegraphics[width=0.29\linewidth]{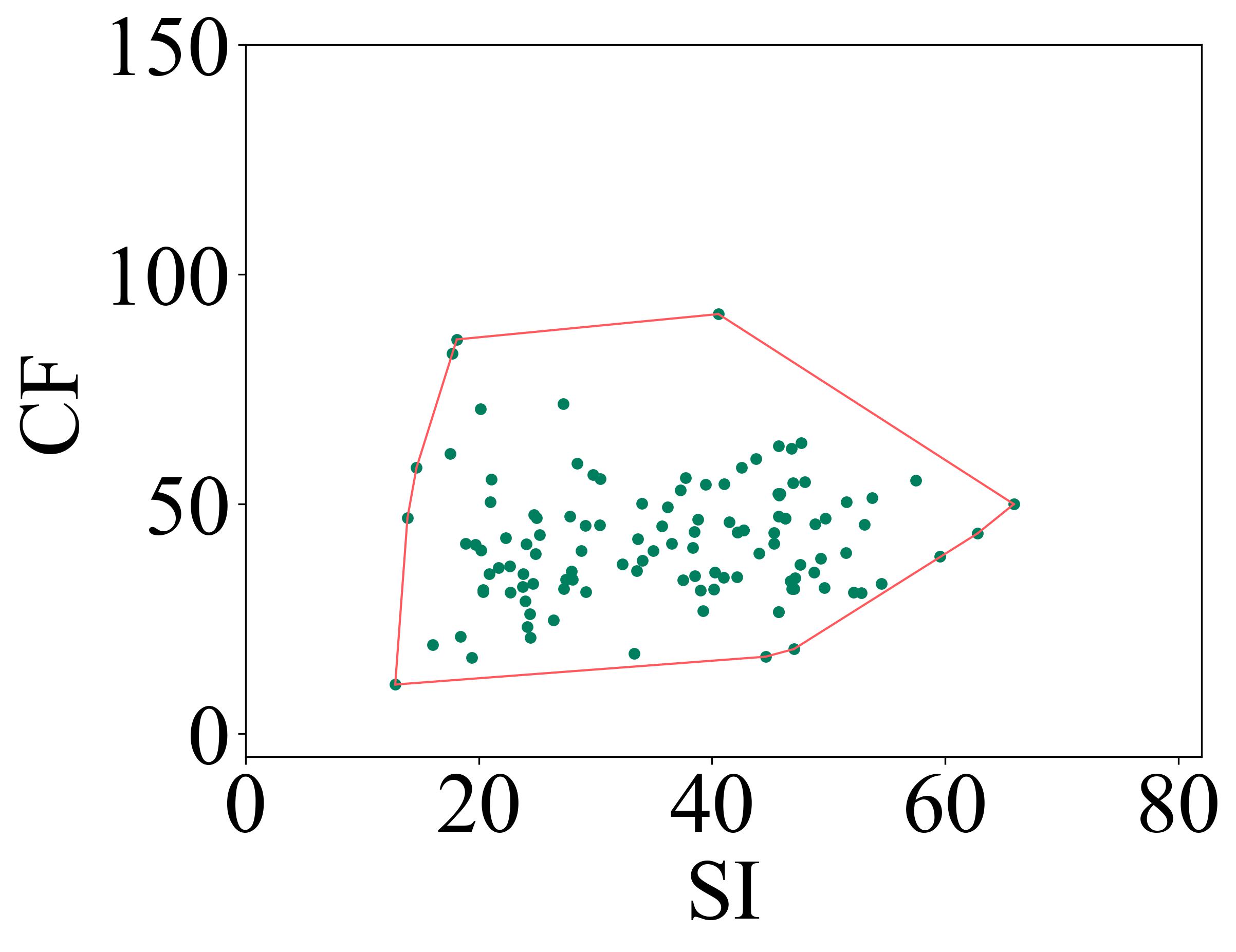}
}
\subfigure[JUFE]{
\includegraphics[width=0.29\linewidth]{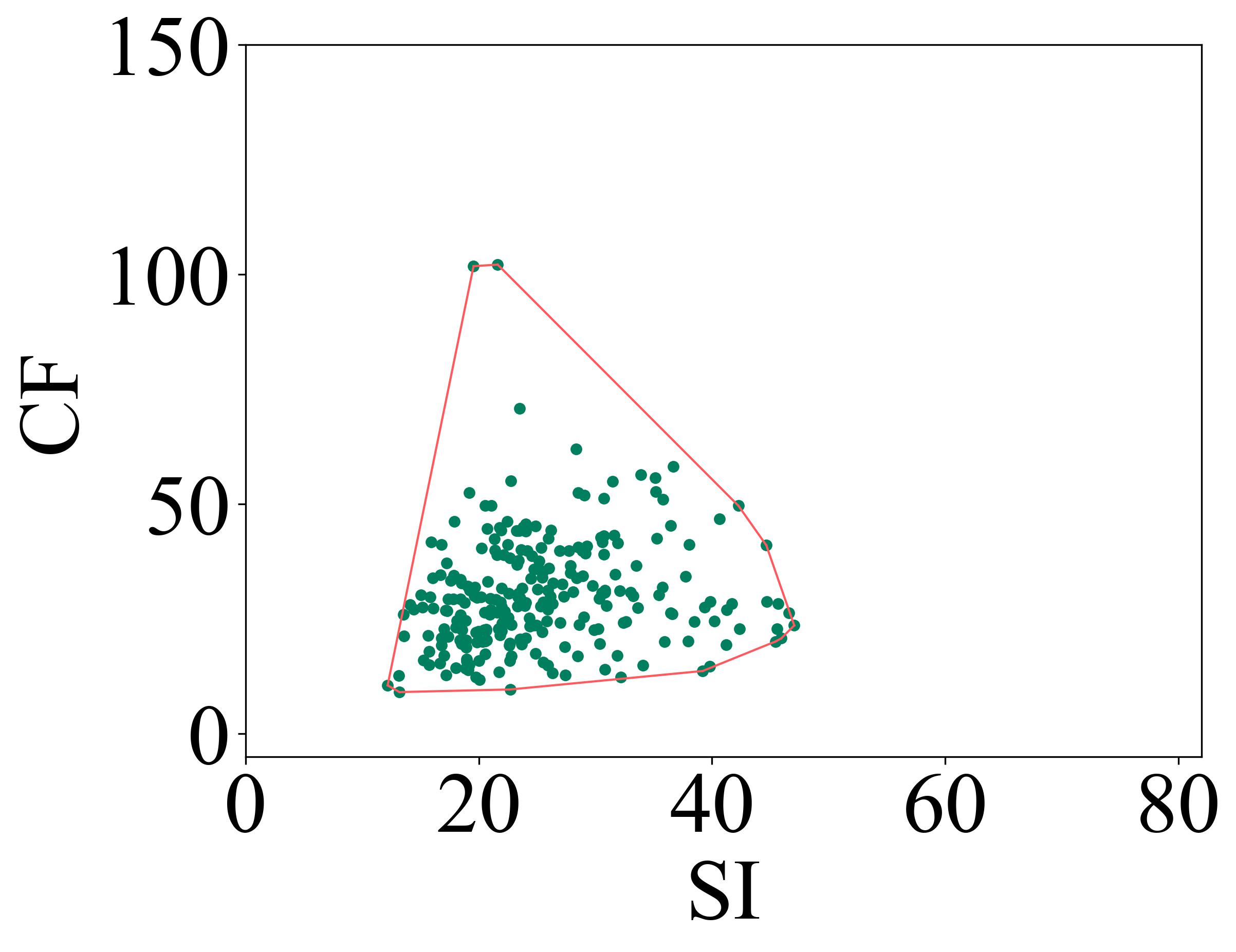}
}
\subfigure[OIQ-10K]{
\includegraphics[width=0.29\linewidth]{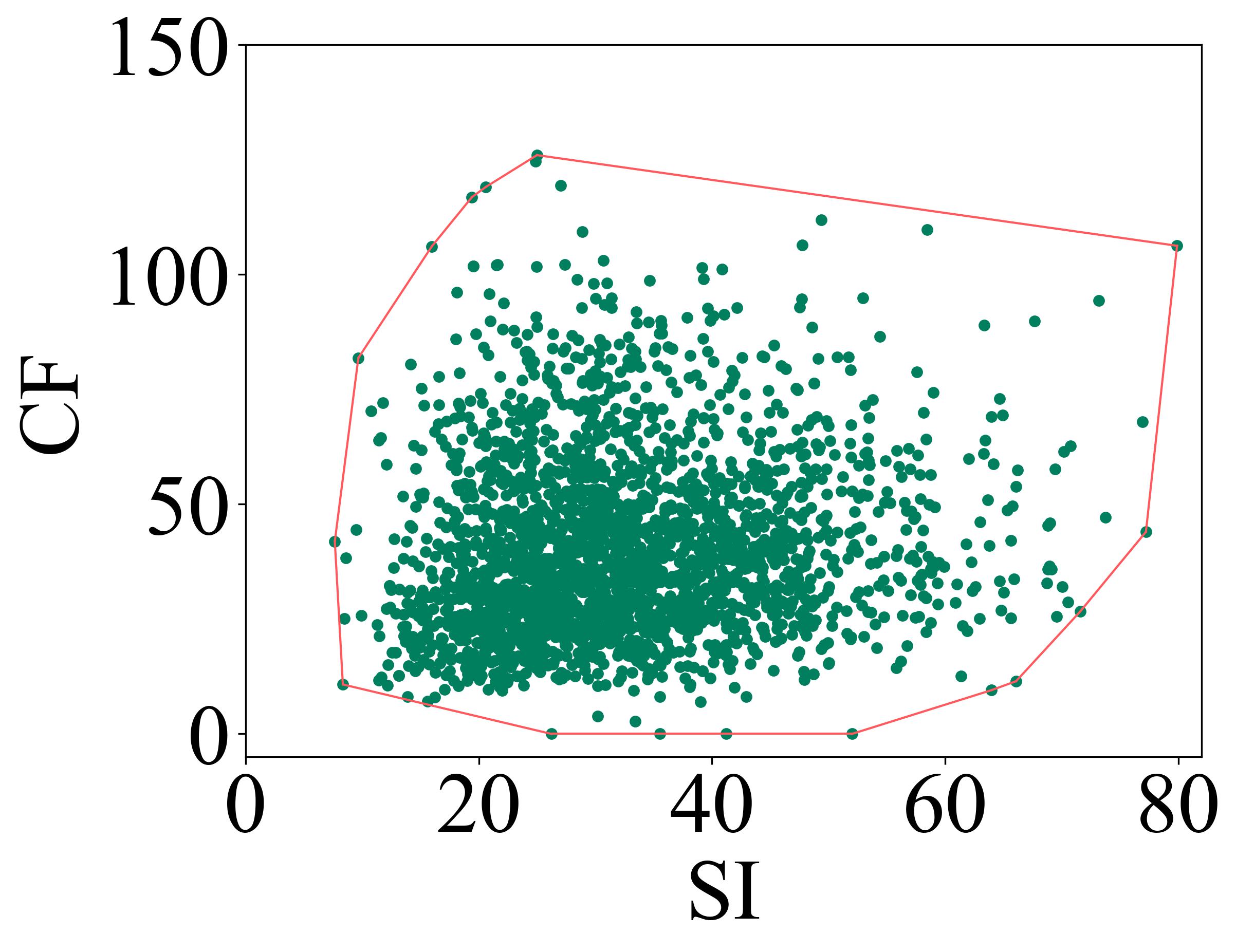}
}
\caption{The scatter plots of Spatial Information (SI) and Colorfulness (CF) of existing OIQA databases and the proposed OIQ-10K database.}
\label{fig:si_cf}
\end{figure}

%------------------------------------------------------------------------
\section{The Proposed OIQ-10K Database}
In this section, we first describe the database construction and the large-scale psychophysical experiment. Then, we give a comprehensive analysis of subjective data.

\begin{figure*}[t]
\centering
\subfigure[Visual examples in the CnoDist situation, without perceptible distortion.]{
\includegraphics[width=0.90\linewidth]{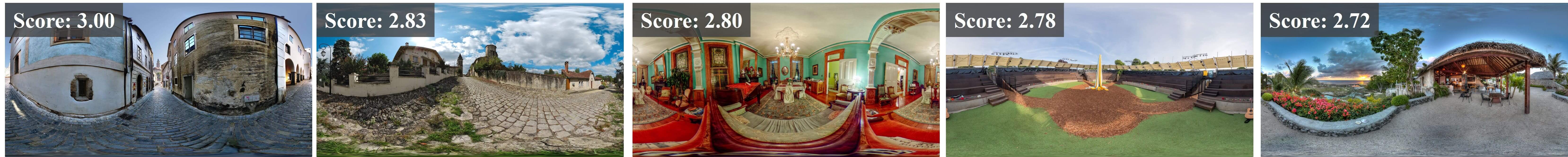}
}
\subfigure[Visual examples in the CdistR1 situation, with one distorted region.]{
\includegraphics[width=0.90\linewidth]{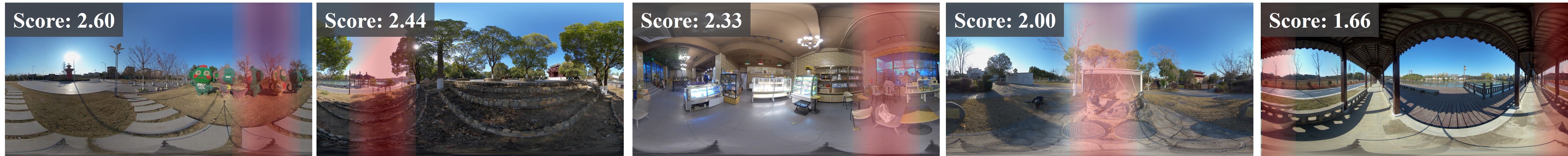}
}
\subfigure[Visual examples in the CdistR2 situation, with two distorted regions.]{
\includegraphics[width=0.90\linewidth]{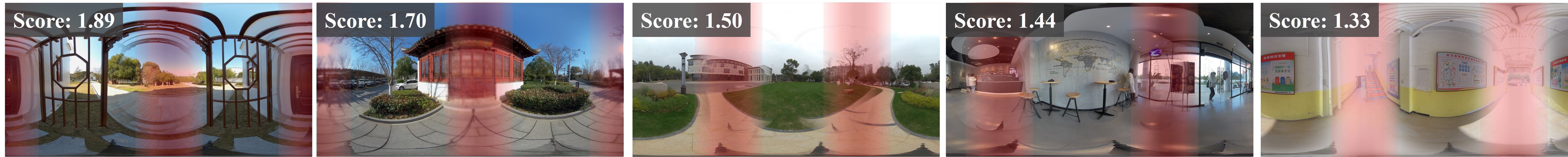}
}
\subfigure[Visual examples in the CdistGl situation, with global distortion.]{
\includegraphics[width=0.90\linewidth]{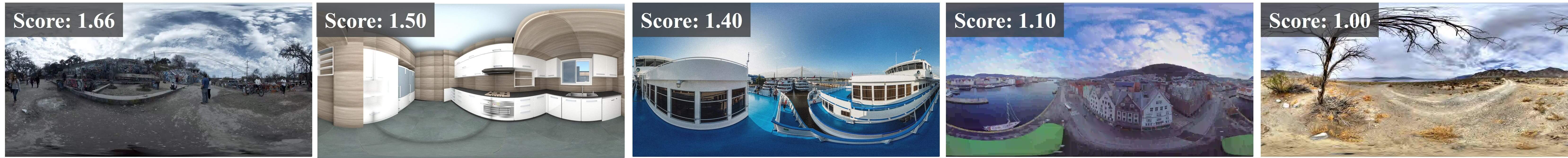}
}
\caption{Visualization of omnidirectional images with different distortion situations in the proposed OIQ-10K database. The distorted region(s) of the visual examples in (b) and (c) are marked in red for better visual presentation.}
\label{fig:database}
\end{figure*}

\subsection{Database Construction}
\label{subsec:da_con}

% Thus, we first classify omnidirectional images into four types according to their distortion ranges, \emph{i.e.}, images without perceptibly distorted region, images with one distorted region, images with two distorted regions, and images with global distortion, which simulates the complicated distortion scenarios in practice. subsequently, we collect omnidirectional images by each type, respectively.

We note that strictly determining whether an image is a reference image is challenging, as it may contain subtle distortion due to complex factors during capture, \emph{e.g.}, particles drifting in the air, slight exposure imbalance and imperfections on the object. Therefore, we subjectively judge whether an omnidirectional image contains distortion. We empirically classify the target omnidirectional images into four categories according to their distortion situations, \ie, images without perceptibly distorted region, images with one distorted region, images with two distorted regions, and images with global distortion, which simulate the complicated distortion scenarios in practice. Among them, the distortion areas of one distorted region and two distorted regions account for one-quarter and half of the entire image, respectively. For clarity, we denote these four situations as ``CnoDist", ``CdistR1", ``CdistR2", and ``CdistGl". Overall, the data collection includes two stages, \ie, the coarse stage and the refinement stage. We first introduce the coarse stage as follows.

\begin{table}[]
\centering
\caption{The established details of the proposed OIQ-10K.}
\label{tab: es_details}
\begin{tabular}{c|c|c}
\toprule
Situation & Coarse Stage   & Refinement Stage \\
\midrule
CnoDist & 1,001+2,498+404=3,903     & 2,500     \\
\midrule
CdistR1 & 258$\times$4$\times$3=3,096   & 209$\times$4$\times$3=2,508     \\
\midrule
CdistR2 & 258$\times$4$\times$3=3,096   & 209$\times$4$\times$3=2,508     \\
\midrule
CdistGl & 2,071+(142+95)+176=2,484 & 2,484     \\
\midrule
Total & 12,579 & 10,000   \\
\bottomrule
\end{tabular}
\end{table}

\textbf{The Coarse Stage.} At this stage, we aim to gather as many images as possible for each situation to fully cover the spatial domain. The details of image collection are as follows.

\begin{itemize} 
 \item \textbf{CnoDist}: 3,903 undistorted images are collected from three sources: 1) 1,001 reference images from the existing OIQA databases ~\cite{sun2018large,duan2018perceptual,chen2019study,qi2020viewport,fang2022perceptual,rai2017dataset,xu2021saliency}; 2) 2,498 images from websites Flickr\footnote{https://www.flickr.com/groups/flickrvr/}; 3) 404 images from the Pixexid\footnote{https://www.pixexid.com}. 
 
 \item \textbf{CdistR1}: 3,096 distorted images are derived from expanding the JUFE database \cite{fang2022perceptual}. For each distorted images generated from 258 reference images, the distortion level for each of the four distortion types is extended from one of three levels to all three levels.
 
 \item \textbf{CdistR2}: 3,096 distorted images are obtained through expanding each image in \textbf{CdistR1}, where one distorted region is transformed into two distorted regions by adding the same distortion of the original fisheye image to another non-adjacent fisheye image.
  
 \item \textbf{CdistGl}: 2,484 distorted images are collected from three sources: 1) 2,071 distorted images from these publicly available OIQA databases \cite{sun2018large,duan2018perceptual,chen2019study,cao2021quality,qi2020viewport,huang2018modeling}; 2) 237 images from Flickr, comprising 142 images with artificially added JPEG compression and 95 images with Gaussian noise, each type featuring three levels of distortion; 3) 176 globally authentic distorted images from Pixexid.
\end{itemize}

Finally, we can obtain 12,579 omnidirectional images with four distortion situations in this stage. The details are summarized in Table~\ref{tab: es_details}.

\textbf{The Refinement Stage.} This stage aims to exclude the images with repeated content, meaningless symbols, and other flaws, while maintaining each situation contains around 2,500 images. For the collected 3,903 images in CnoDist, we first exclude the \textit{repeatability}. We represent each image as a 1$\times$512 vector by extracting features from the layer just before the fully connected layer of a pre-trained ResNet18~\cite{he2016deep}, and exclude the images with same features. Then the database shaping technique \cite{vonikakis2017probabilistic} is adopted to analyze these features, exclude images with high correlation to arrive at a compact but attribute-wise balanced database. Finally, we manually select the remaining images and obtain 2,500 images.

% , including 115 images from JUFE~\cite{fang2022perceptual}, 10 images from CVIQ~\cite{sun2018large}, 10 images from OIQA~\cite{duan2018perceptual}, 60 images from Salient360!~\cite{rai2017dataset}, 436 images from Xu~\textit{te al.}~\cite{xu2021saliency}, 7 images from NBU-SOID~\cite{qi2020viewport}, 12 images from LIVE 3D VR IQA~\cite{chen2019study}, 150 images from Pixexid, and 1,700 images from Flickr. 
For the 3,096 images in CdistR1 and 3,096 images in CdistR2, we meticulously remove simple-content images from the 258 reference images, resulting in 209 reference images and corresponding 2,508 distorted images, with this process carried out simultaneously for both situations. For the omnidirectional images with global distortion. we keep all of them. Ultimately, we collect 10,000 omnidirectional images with four distortion situations. Figure~\ref{fig:si_cf} shows the diversity of our proposed database in terms of spatial information~\cite{yu2013image} and colorfulness~\cite{hasler2003measuring}.

\subsection{Large-scale Psychophysical Experiment}
\label{subsec:sub_eva}

% Our goal is to guide service providers in quickly screening images based on user acceptability, which does not rely upon precise quality scores. Thus,

\textbf{Environmental Settings.} As suggested in the recommendations ITU-R BT.500-13~\cite{bt2002methodology}, we adopt the single-stimulus (SS) method to collect subjective annotations. Note that we refer to ITU-R BT.500-13 (which focuses on evaluating the quality of television images) rather than ITU-T Rec. P.919 \cite{recommendation2020subjective} (which is centered on 360º videos), since the former's settings are more suitable for omnidirectional images and can offer more detailed descriptions about the single-stimulus method and the data screening method, while the latter primarily focuses on providing settings for head-mounted displays (HMDs) and has also been adopted in existing OIQA studies~\cite{sun2018large,duan2018perceptual,chen2019study}. Inspired by the previous work~\cite{yan2022subjective,li2019accann}, we set three labels to represent the quality of omnidirectional images: “Good”, “Fair”, and “Poor”, whose corresponding scores are 3, 2, and 1, respectively. We considered two factors. First, we aim to provide service providers and users with a clear understanding of the quality of omnidirectional images, helping them quickly and automatically decide whether to retain or discard the images. A more refined measurement approach yields limited benefit to the results. Second, subjects prefer this approach due to its reduced cognitive load.
% Similar to our previous study \cite{yan2022subjective}, we set three labels to represent the quality of omnidirectional images: ``Good", ``Fair", and ``Poor", whose corresponding scores are 3, 2, and 1, respectively. This methodology requires much less cognitive load and collects more reliable data~\cite{li2019accann}.
An in-lab subjective experiment is conducted carefully. The subjects can choose to sit on a swivel chair or stand freely in an empty and quiet room. To streamline the entire annotation process, we develop an experimental platform utilizing Unity3D software, each image is displayed for 15 seconds, and the screen resolution of images is uniformly set to 7,680 $\times$ 4,320. Subjects utilize a set of HTC VIVE equipment to label each image in the annotation interface after viewing images, and the annotation data including quality scores, head movement records and eye movement records are automatically stored. Note that the labels for local and global distortions are automatically determined in the coarse stage. Approximately 160 participants, consisting of 120 males and 40 females, are invited to participate in the subjective experiment. Their ages range from 18 to 34, and all participants have unimpaired visual perception.  

\textbf{Annotation Process.} We divide the 10,000 images into 100 groups and ensure that each group receives annotations from 5 different annotators. Subsequently, we calculate the quality variance for each image, and the images with a variance smaller than the first 25\% quartile of quality distribution are deemed relatively uncontroversial on perceptual quality. Then, the remaining images (deemed as ``uncertain" samples) are divided into 58 groups, and each group receives annotations from 5 different subjects. Before starting evaluation, the whole procedure is explained to each subject, and 10 test samples are arranged to help subjects familiarize with this task. Note that the subjects are allowed to rest whenever they feel fatigued. Finally, we collected more than 87,000 scores from 160 subjects. The entire annotation process takes approximately eight months. After removing unreliable subjects~\cite{bt2002methodology}, we compute the mean opinion score (MOS) of each omnidirectional image. Some visual samples with different degradation situations are shown in Figure~\ref{fig:database}.

\begin{figure}[t]
\centering
\subfigure[]{
\includegraphics[width=0.46\linewidth]{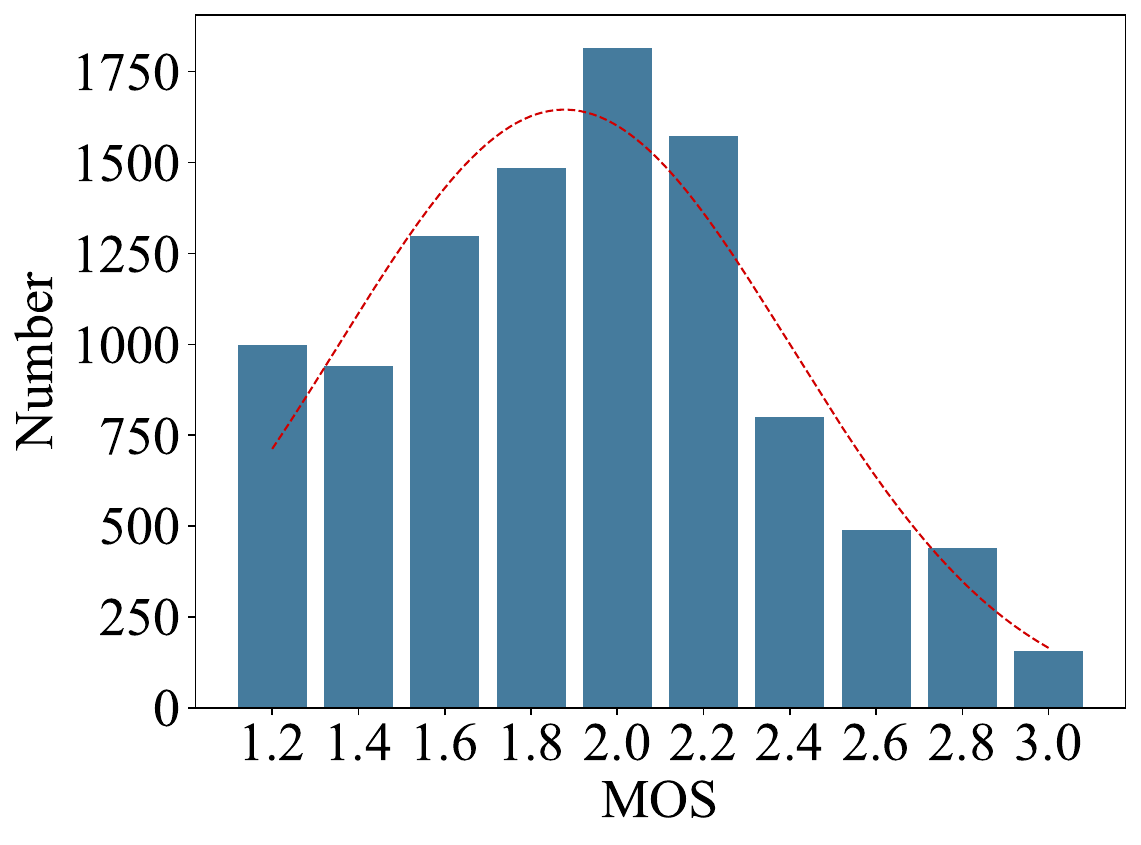}
}
\subfigure[]{
\includegraphics[width=0.46\linewidth]{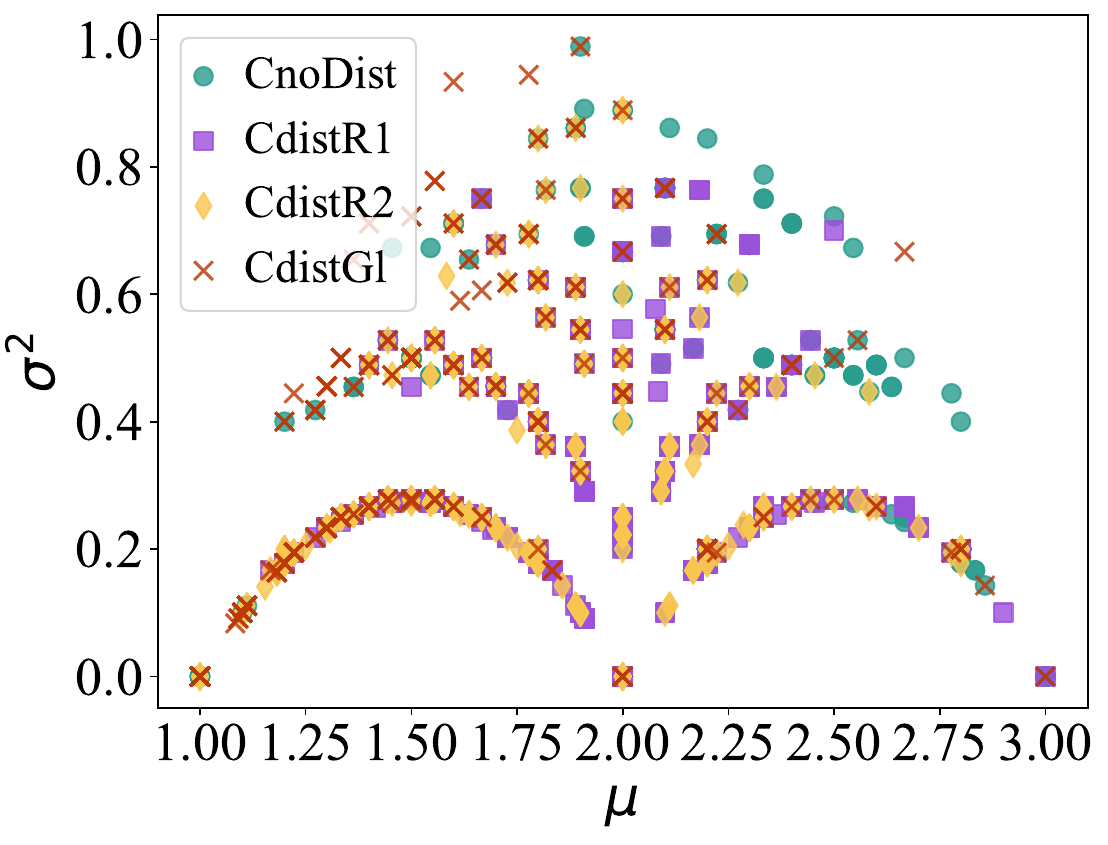}
}
\subfigure[]{
\includegraphics[width=0.95\linewidth]{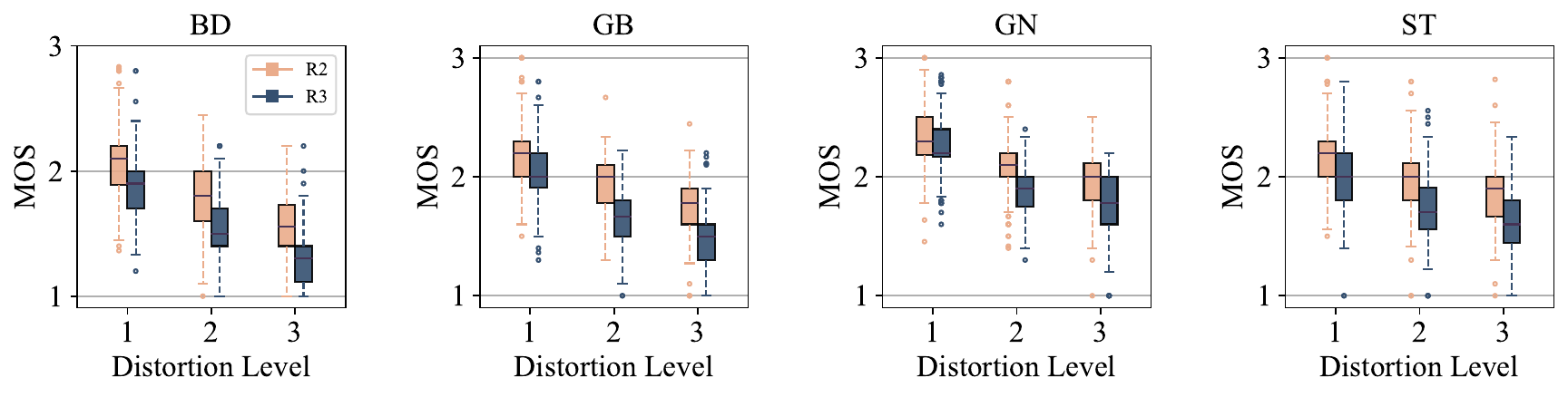}
}
\caption{Statistics in OIQ-10K database: (a) The distribution of MOSs; (b) The scatter plot of the mean ($\mu$) and variance ($\sigma^2$ ) of the images with different degradation situations; (c) The box plot of MOSs in the CdistR1 and CdistR2 situations.}
\label{fig:mos_analysis}
\end{figure}

\begin{figure}[t]
\centering
\subfigure[The starting point.]{
\includegraphics[width=0.95\linewidth]{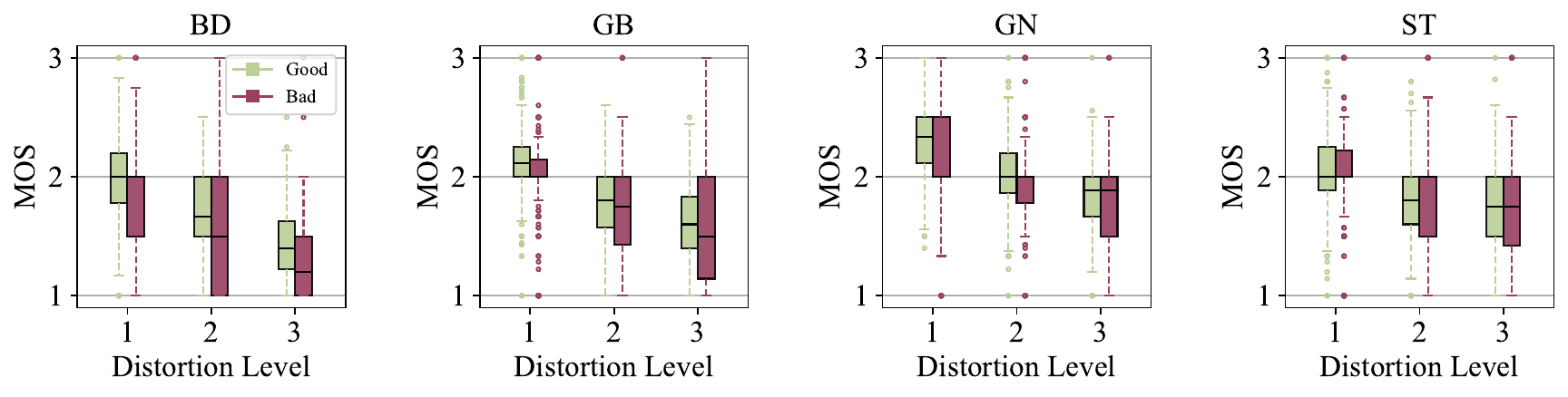}
}
\subfigure[The ending point.]{
\includegraphics[width=0.95\linewidth]{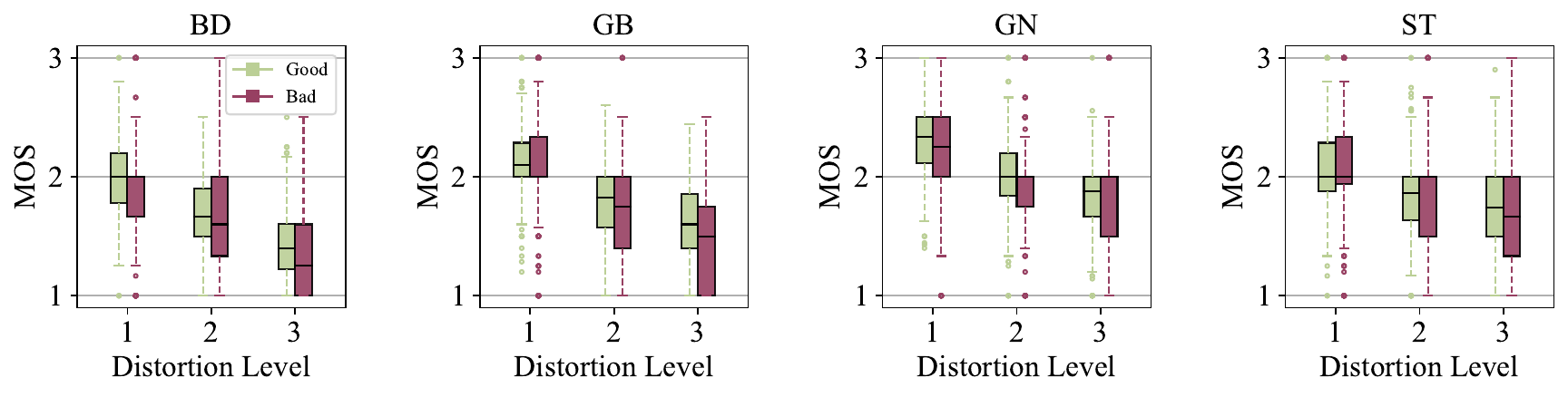}
}
\caption{The box plot of MOSs under different viewing conditions.}
\label{fig:start_point}
\end{figure}

\begin{table}[t]
  \centering
  \caption{Impact of viewing condition. The viewing points (\textit{i.e.}, the starting and ending points) have a significant impact (ANOVA with p$<$0.05) on the MOS.}
  \label{tab:viewing_condition}
  \begin{tabular}{cccc}
 \toprule
 \# & Correlation to MOS & PLCC & SRCC \\
 \midrule
 1 & good starting point & 0.9211 & 0.9299 \\
 2 & bad starting point & 0.6883 & 0.7136 \\
 3 & good ending point & 0.9338 & 0.9343 \\
 4 & bad ending point & 0.7168 & 0.7348 \\
 \bottomrule
  \end{tabular}
\end{table}

\subsection{Analysis of Subjective Data}

The distribution of subjective quality scores is illustrated in Figure~\ref{fig:mos_analysis} (a), where we can observe that the distribution is nearly normal and the scores cover all ranges. We also show the scatter plot of the mean and variance of quality scores in Figure~\ref{fig:mos_analysis} (b). We find that the proportion of low scores increases with the expansion of distorted regions, which is aligned with human perception. Besides, subjects hold varied judgments on the images with middle quality, this phenomenon also emerges in other IQA databases as depicted  in~\cite{zhang2021uncertainty,talebi2018nima}. To investigate the influence of distortion situation, we qualitatively analyze the box plot of MOSs under different viewing conditions on the CdistR1 and CdistR2 situations. As shown in Figure~\ref{fig:mos_analysis} (c), we can observe that the MOSs of the images in the CdistR1 situation are higher than that in the CdistR2 situation regardless of distortion types and levels. This phenomenon is remarkable when images suffer from BD distortion.

In addition, we continue to explore viewing conditions along the lines of the previous work~\cite{fang2022perceptual} (\ie, the good starting point and the bad starting point). Note that a good starting point indicates that the initial viewing region seen by subjects is of high quality, and vice versa. Inspired by the recency effect~\cite{hands2001recency}, we also study the impact of the ending point (\ie, the last observed region). For Figure~\ref{fig:start_point} (a), we have several interesting findings. First, the MOSs of images viewed from a good starting point are slightly higher than those from a bad starting point, indicating a pleasant first impression brings a positive assessment. Second, the MOS distribution ranges of the images viewed from a bad starting point are wider than that of a good starting point, the reason may be that an unpleasant first impression affects the subsequent experience constantly and leads to a vacillating evaluation. We then study the impact of the ending point, and the results are shown in Figure~\ref{fig:start_point} (b). The observation is similar to Figure~\ref{fig:start_point} (a). In addition, we quantitatively analyze the correlation between the overall MOSs, and MOSs in different viewing conditions and the results are shown in Table~\ref{tab:viewing_condition}. The MOSs with a great initial impression are more relevant to the overall MOSs than the MOSs with an inferior impression, this situation also happens in the ending point. It means that a good beginning or ending promotes the stability of subjective quality assessment.

\begin{figure}[t]
\centering
\includegraphics[width=1\linewidth]{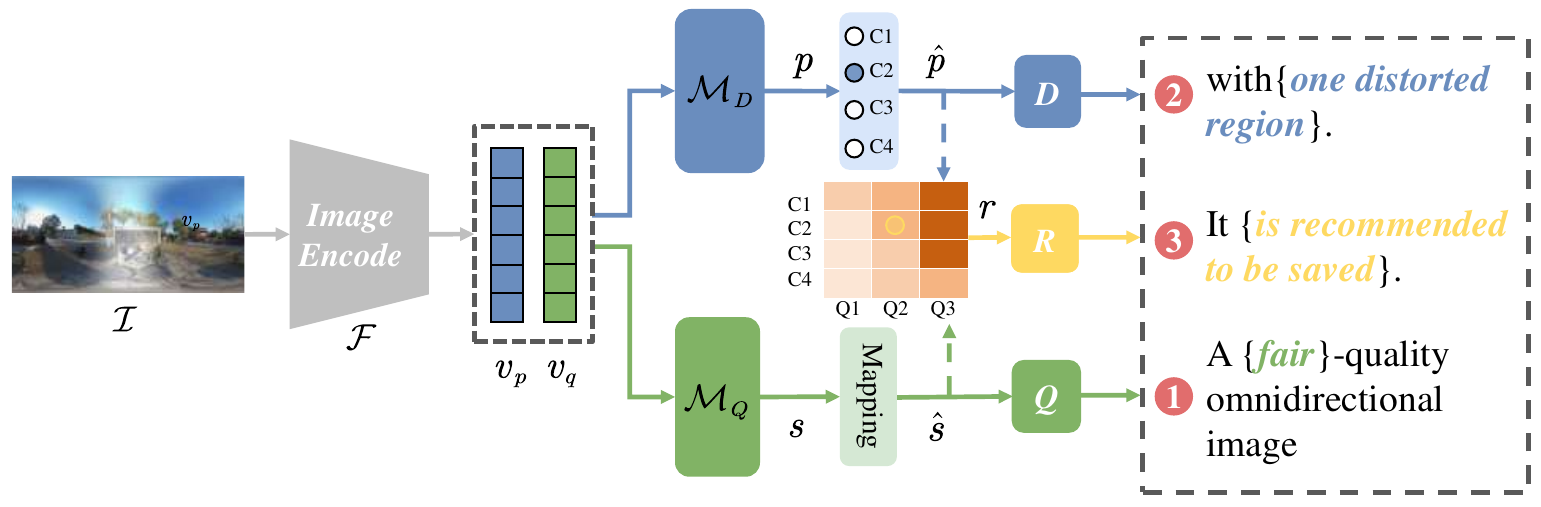}
\caption{The flowchart of generating quality captions for omnidirectional images. The caption \textit{D} is determined by the distortion situation prediction network; the caption \textit{Q} is obtained according to the result of the quality score prediction network; the caption \textit{R} is determined by the outputs of these two networks jointly. Q1, Q2, and Q3 denote the ``poor", ``fair", and ``good", respectively. C1, C2, C3 and C4 denote the ``CnoDist", ``CdistR1", ``CdistR2", and ``CdistGl", respectively. For concise illustration, we simplify the design of other modules of the proposed model.}
\label{fig:cgp}
\end{figure}

\begin{figure*}[t]
\centering
\includegraphics[width=0.8\linewidth]{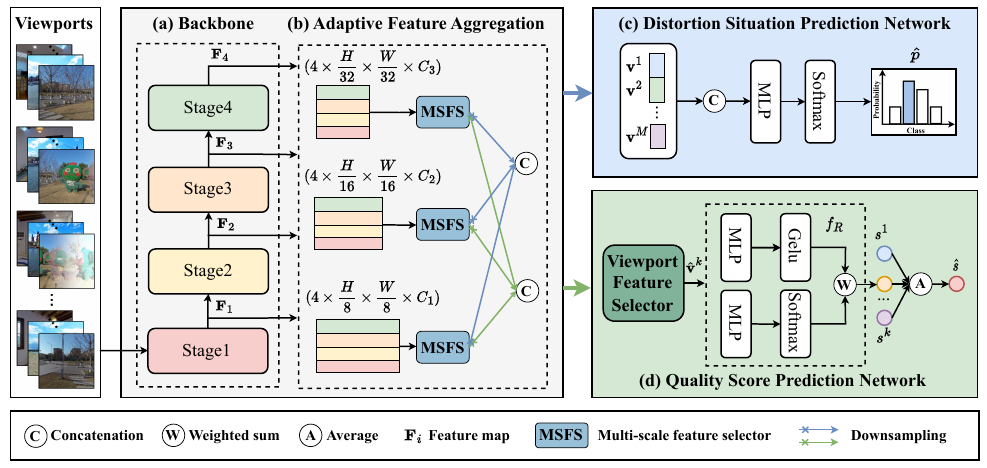}
\caption{The architecture of the proposed IQCaption. It contains four parts: (a) backbone, (b) adaptive feature aggregation module, (c) distortion situation prediction network, and (d) quality score prediction network.}
\label{fig:framework}
\end{figure*}

%-------------------------------------------------------------------------
\section{The Proposed IQCaption360 Model}
\label{sec:propsed_model}

In this section, we first illustrate the process of quality caption generation in our work. Subsequently, a detailed introduction about the proposed IQCaption360 is presented.

\subsection{Quality Caption Generation}
\label{subsec:gqc}
Different from previous works that use numerical results to describe the quality of omnidirectional images, we aim to generate quality-aware textual descriptions. Inspired by these advanced studies~\cite{yang2022fine,zhang2023blind}, the quality captions are generated in a manner of textual template in this work, and the generation process in our proposed model is shown in Figure~\ref{fig:cgp}. Differently from ~\cite{yang2022fine}, we treat quality prediction as a primary regression task that allows for learning in a more precise manner, with an auxiliary classification task to enhance performance. Specifically, given an input omnidirectional image $\mathcal I$, the image embedding $v_p$ and $v_q$ can be obtained using the image encoder $\mathcal{F}$. The distortion situation prediction network $\mathcal M_D$ and the quality score prediction network $\mathcal M_Q$ are designed to decoder the image embeddings to corresponding results, which can be formulated as:
\begin{align}
\label{eq:qcr}
    \begin{split}
    \{v_p,v_q\}=\mathcal{F}(\mathcal{I};\theta_f),\\
    p=\mathcal{M}_D(v_p;\theta_D),
    s=\mathcal{M}_Q(v_q;\theta_Q),
    \end{split}
\end{align}
where $p$ is the predicted probability of distortion situation; $s$ is the predicted quality score; $\theta_f$, $\theta_D$ and $\theta_Q$ are the parameters of $\mathcal{F}$, $\mathcal{M}_D$ and $\mathcal{M}_Q$, respectively. To generate quality captions, we first convert the numerical results to textual results. For distortion situation prediction, $\hat p$=argmax($p$), where $\hat p\in D=$ \{``no perceptibly distorted region", ``one distorted region", ``two distorted regions", ``global distortion"\}. For quality prediction, we use value-to-text mapping to convert the numerical $s$ to $\hat s\in Q=$\{``good", ``fair", ``poor"\}. For the recommendation $r\in R=$\{``should be saved", ``is recommended to be saved", ``is recommended to be discarded", ``should be discarded"\}, we introduce a recommendation table to determine the $r$ of each image. Specifically, the recommendation is decided by the mapped quality level $\hat s$ and distortion situation $\hat p$ and has $3\times4=12$ outcomes according to the combination of these two factors, and the outcomes are further classified into four types of recommendations. Similar to~\cite{zhang2023blind}, we define a textual template to combine the results together: ``\emph{A} \{$\hat s$\}\emph{-quality omnidirectional image with} \{$\hat p$\}. \emph{It} \{$r$\}".

\subsection{The Proposed IQCaption360}
The detailed architecture of the proposed IQCaption360 is shown in Figure~\ref{fig:framework}. We first extract a viewport sequence from an omnidirectional image, and a backbone is applied to extract multi-scale features of each viewport. Then the adaptive feature aggregation (AFA) module is designed for tailoring the general multi-scale features to task-specific features. Subsequently, we design a distortion situation prediction network (DSPN) and a quality score prediction network (QSPN), whose inputs are task-specific features output by the AFA module. Finally, the quality caption can be obtained by combining the outputs of these two networks. The details are as follows.

\textbf{Viewport Feature Extraction.} Considering that the subjective quality of omnidirectional images highly depends on viewports~\cite{li2019viewport} and most scene information is concentrated in the low latitude regions as well as being visited more frequently than other parts~\cite{zhang2022no}, we select the viewports equidistant along the equator of omnidirectional image. To effectively capture the global quality and the cues of the distortion situation, the extracted viewport sequence covers the entire equatorial region. Given $N$ omnidirectional images $\mathcal{O}=\{\mathcal{I}_n\}_{n=1}^N$, the viewports and their coordinate in $n$-th image $\mathcal{I}_n$ is $\{{\mathcal{V}_n^m}, (\phi_n^m,0)\}_{m=1}^M$, where $M$ is the number of viewports and $\phi_n^{m}=\phi_n^{m-1}+A$ ($A$ is a constant).

For a given image $\mathcal{I}$ ($n$ is omitted for simplicity) and its viewports $\{\mathcal{V}^m\}_{m=1}^M$, we first apply two 3$\times$3 convolutional layers to compress the shape of each viewport from $(H, W, 3)$ to $(\frac{H}{4}, \frac{W}{4}, C)$ for improving computational efficiency. Then, we adopt the neighborhood attention transformer (NAT) \cite{hassani2023neighborhood} as the backbone, which introduces local inductive biases and maintains translational equivariance, while enjoying a linear time and space complexity. We follow the design norm of Swin Transformer~\cite{liu2021swin}, the backbone consists of four stages. The number of NATs in each stage is 2, 2, 5, and 3. Afterward, the multi-scale features of each viewport can be obtained by concatenating the outputs of four stages, whose size are $(\frac{H}{8}, \frac{W}{8}, C_1)$, $(\frac{H}{16}, \frac{W}{16}, C_2)$, $(\frac{H}{32}, \frac{W}{32}, C_3)$ and $(\frac{H}{32}, \frac{W}{32}, C_4)$. For the $m$-th viewport, the above procedure can be formulated as:
\begin{align}
\label{eq:vp_ext}
    \textbf{F}^m=f_{nat}(\mathcal{V}^m;\theta_{nat}),
\end{align}
where $\textbf{F}^m=\{\textbf F_{\bar{s}}^m\}_{\bar{s}=1}^{\bar{S}}$ and $\textbf F_{\bar{s}}^m$ denotes the features from the $\bar{s}$-th stage of backbone; $\bar{S}=4$ denotes the number of stages; $f_{nat}$ and $\theta_{nat}$ denote the backbone and its parameters, respectively.

\textbf{Adaptive Feature Aggregation Module.} Inspired by the fact that different visual tasks rely on specific features~\cite{yang2022fine,liu2020dynamic,yang2021image}, we design an AFA module for tailoring the general multi-scale features to task-specific features. For the multi-scale features $\textbf{F}^m$ of the $m$-th viewport, we first use four $1\times1$ convolution layers and the bilinear interpolation operation to unify feature shapes. We repeat this procedure at different scales to reduce the information loss in the reshaping process, which can be formulated as:
\begin{align}
\label{eq:alter_shape}
    \bar{\textbf{F}}_{\bar{s}}^m=\Psi_{\bar{s}} (\textbf{F}^m),
\end{align}
where $\Psi_{\bar{s}}$ denotes the bilinear interpolation function, which interpolates the input $\textbf{F}^m$ to $\bar{\textbf{F}}_{\bar{s}}^m\in \mathbb{R}^{\bar S\times H_{\bar{s}}\times W_{\bar{s}}\times C_{\bar{s}}}$ with a unified shape, and the shape of features from the $\bar{s}$-th stage is $(H_{\bar{s}}, W_{\bar{s}}, C_{\bar{s}})$. Finally, three multi-scale features are obtained with size of $(4, \frac{H}{8}, \frac{W}{8}, C_2)$, $(4, \frac{H}{16}, \frac{W}{16}, C_1)$ and $(4, \frac{H}{32}, \frac{W}{32}, C_3)$. Then we design a multi-scale feature selector (MSFS) based on the DFIM \cite{liu2020dynamic} to obtain the task-specific features:
\begin{align}
\label{eq:tailor_feats}
    \bar{\textbf{F}}_{task,\bar{s}}^m=f_{msfs}(\bar{\textbf{F}}_{\bar{s}}^m;\theta_{msfs}),
\end{align}
where $f_{msfs}$ denotes the MSFS module, and $\theta_{msfs}$ denotes its parameters; $\bar{\textbf{F}}_{task,\bar{s}}^m\in \mathbb{R}^{H_{\bar{s}}\times W_{\bar{s}}\times C_{\bar{s}}}$ denotes the fused features at the $\bar s$-th scale. Subsequently, we apply a layer normalization (LN) and the pooling operation on the fused features to obtain the feature vector $\textbf{v}_{task,s}^m\in \mathbb{R}^{B\times C_{\bar s}}$, where $B$ denotes the batch size. The final $\textbf{v}_{task}^m$ across multiple scales can be obtained by concatenation operation:
\begin{align}
\label{eq:concat_feats}
    \textbf{v}_{task}^m=\textbf{v}_{task,1}^m\cup \textbf{v}_{task,2}^m\cup \cdots \cup\textbf{v}_{task,\bar S}^m,
\end{align}
where $\textbf{v}_{task}^m\in \mathbb{R}^{B\times \sum_{\bar{s}=1}^{\bar S} C_{\bar s}}$ denotes the specific quality-aware feature vector of the $m$-th viewport. Here, two kinds of vectors are obtained, $\textbf{v}_{dspn}^m$ for the distortion situation prediction task, and $\textbf{v}_{qspn}^m$ for the quality score prediction task.

\begin{figure}[t]
\centering
\includegraphics[width=0.9\linewidth]{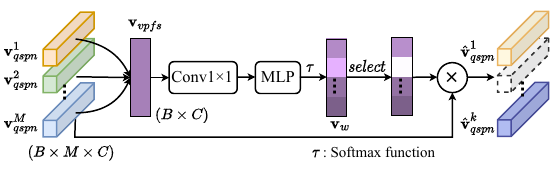}
\caption{The architecture of VPFS.}
\label{fig:vpfs}
\end{figure}

% Besides, we find that experience consistency is of great importance in the process of viewing omnidirectional images. 

% In this paper, we design a distortion range prediction network (dspn) to capture the experience consistency. 

\textbf{Distortion Situation Prediction Network.} Since utilizing the auxiliary task facilitates performance improvement~\cite{zhang2023blind,ma2017end,zhou2021omnidirectional}, we first concatenate the vectors $\textbf{v}_{dspn}^m$ of all viewports from the same images about this task:
\begin{align}
\label{eq:dspn_feats}
    \textbf{v}_{dspn}=\textbf{v}_{dspn}^1\cup \textbf{v}_{dspn}^2\cup \cdots \cup\textbf{v}_{dspn}^M,
\end{align}
where $\textbf{v}_{dspn}\in \mathbb{R}^{B\times M\sum_{\bar s=1}^{\bar S} C_{\bar s}}$ denotes the DSPN-related features of entire omnidirectional image. Then the predicted probability of distortion situation of the $n$-th omnidirectional image can be obtained by applying MLP layer and Softmax function:
\begin{align}
\label{eq:dspn_prob}
    \hat{p}_n=\text{Softmax}(\text{MLP}(\textbf{v}_{dspn})),
\end{align}
where $\hat{p}_n=\{\hat{p}^1_n,\hat{p}^2_n,\cdots,\hat{p}^{\bar D}_n\}$, and $\bar D$ equates to 4.

\textbf{Quality Score Prediction Network.} We consider the experience inconsistency in the heterogeneously distorted omnidirectional image, \ie, viewports from the same image may differ from each other notably in quality and content~\cite{fang2022perceptual}. To address this issue, we introduce a viewport feature selector (VPFS) module in the regression phase. The VPFS module dynamically assigns weights to viewports and discards irrelevant ones, whose architecture is illustrated in Figure~\ref{fig:vpfs}. Specifically, we first obtain the merged vector $\textbf{v}_{vpfs}$ by the addition operation:
\begin{align}
\label{eq:vpfs_add}
    \textbf{v}_{vpfs}=\textbf{v}_{qspn}^1+\textbf{v}_{qspn}^2+\cdots +\textbf{v}_{qspn}^M,
\end{align}
where $\textbf{v}_{vpfs}\in \mathbb{R}^{B\times \sum_{\bar s=1}^{\bar S} C_{\bar s}}$ denotes the QSPN-related features. Then we use a 1$\times$1 convolutional layer to merge the information of $\textbf{v}_{vpfs}$ in the channel dimension, and an MLP layer to map the merge vector to a weight vector $\textbf{v}_{w}$ after the convolution operation:
\begin{align}
\label{eq:vpfs_fc}
    \textbf{v}_{w}=\text{Softmax}(\text{MLP}(\text{Conv}(\textbf{v}_{vpfs}))),
\end{align}
where $\textbf{v}_{w}=\{w^1,w^2,\cdots,w^M\}$. We designate a parameter $K$ to choose the top-$K$ $\textbf{v}_{vpfs}^i$ with the highest weight:
\begin{align}
\label{eq:vpfs_w}
    \hat{\textbf{v}}^m_{qspn}=w^m\odot \textbf{v}_{qspn}^m,m=1,2,\cdots,M,
\end{align}
where $\odot$ denotes the element-wise multiplication; $\hat{\textbf{v}}^m_{qspn}$ denotes the weighted QSPN-related features of the $m$-th viewport. Note that $\sum_m \mathbb I(w^m=0)=M-K$, and $\mathbb I(\cdot)$ denotes the indicator function. Finally, the predicted quality score of the $n$-th omnidirectional image can be obtained:
\begin{align}
\label{eq:vpfs_s}
    \begin{split}
    \hat{s}_n^k=f_R(\hat{\textbf{v}}^k_{n,qspn};\theta_R), \\
    \hat{s}_n=\frac{1}{k}\sum_{k=1}^K \hat{s}_n^k,
    \end{split}
\end{align}
where $\hat{s}_n^k$ denotes the predicted quality score of $k$-th viewport of $n$-th omnidirectional image; $f_R$ and $\theta_R$ represent the regression module and its parameters, respectively.

\textbf{Loss Function for Multitask Learning.} During training, we sample a minibatch $\mathcal{B}$ from $\mathcal{O}$ in each iteration, and the cross-entropy loss is adopted to optimize the DSPN module:
\begin{align}
\label{eq:loss_dspn}
    \mathcal{L}_{dspn}(\textbf{p},\hat{\textbf{p}})=-\frac{1}{N}\sum_{n=1}^N p_n log(\hat{p}_n)+(1-p_n)log(1-\hat{p}_n),
\end{align}
where $\textbf{p}$ and $\hat{\textbf{p}}$ denote the ground-truth probabilities and predicted probabilities from $\mathcal{B}$, respectively. 

Meanwhile, we use the Norm-in-Norm loss \cite{li2020norm} to optimize the QSPN module:
\begin{align}
\label{eq:loss_qspn}
    \mathcal{L}_{qspn}(\textbf{s},\hat{\textbf{s}})=\frac{1}{\varepsilon N}\sum_{n=1}^N||s_n-{\hat s}_n||^{\gamma},
\end{align}
where $\textbf{s}$ and $\hat{\textbf{s}}$ denote the subjective quality score and predicted quality score from $\mathcal{B}$; $\varepsilon$ is a normalization factor; $\gamma$ is a hyperparameter.

Similar to the training strategy of multi-task network \cite{zhang2023blind}, we use the dynamic weight average method \cite{liu2019end} to automatically assign weights to each task:
\begin{align}
\label{eq:loss_total}
    \mathcal{L}_{total}=\lambda_1 \mathcal{L}_{dspn}+\lambda_2 \mathcal{L}_{qspn},
\end{align}
where the weighting $\lambda_k$ for task $k$ is defined:
\begin{align}
\label{eq:loss_qspn}
    \lambda_k(t)=\frac{\text{exp}(w_k(t-1)/T)}{\sum_i\text{exp}(w_i(t-1)/T)},
    w_k(t-1)=\frac{\mathcal{L}_k(t-1)}{\mathcal{L}_k(t-2)}.
\end{align}
where $w_k$ calculates the relative descending rate; $t$ is an iteration index; $T$ denotes a parameter that controls the softness of task weighting. Note that $\lambda_k$ approaches 1 with $T$ increasing.

\section{Quantitative Experiments}
\label{sec:qu_ex}

\subsection{Experimental Settings}
\label{subsec:ex_se}

\textbf{Experiment Methodology.} We randomly split the omnidirectional images in the OIQ-10K database into 80\% and 20\% for training and testing, respectively. We extract eight viewports from the equator of each omnidirectional image and the longitude offset $A$ is 45°. The size of input viewport is set to 224$\times$224$\times$3. The adopted backbone, \ie, NAT~\cite{hassani2023neighborhood}, is pre-trained on the ImageNet~\cite{russakovsky2015imagenet}, and the proposed model is implemented by the PyTorch framework \cite{paszke2017automatic}. The parameters of our model are optimized by the Adam \cite{kingma2014adam} with an initial learning rate of $10^{-4}$ and the cosine decay learning rate with the minimum learning rate of $10^{-6}$. During training, the batch size is set to 32, and the total training process is finished after 50 epochs. All experiments are conducted on the computer with Intel(R) Xeon(R) Gold 6326 CPU@2.90GHz, 24G NVIDIA GeForce RTX A5000 GPU, and 260GB RAM.

\textbf{Performance Evaluation Criteria.} We employ Pearson's linear correlation coefficient (PLCC) and Spearman's rank order correlation coefficient (SRCC) as the criteria to assess the compared models, where a higher value denotes better performance. As suggested in~\cite{sheikh2006statistical}, a five-parameter logistic function is applied before calculating PLCC. Besides, we also use classification accuracy (ACC) to assess the performance of DSPN, which is defined as:

\begin{align}
\label{eq:acc}
    ACC=\frac{N_{acc}}{N_{total}},
\end{align}
where $N_{acc}$ and $N_{total}$ denote the number of correctly classified samples and the total number of samples, respectively.

\begin{table*}[t]
    \centering
    \caption{Performance comparison of state-of-the-art IQA and OIQA metrics on the OIQ-10K database in detail. The best results are marked in \textbf{bold}, and the second best is \underline{underlined}.}
    \label{tab:sota}
    \begin{tabular}{llcccccccccc}
        \toprule
        \multirow{2}*{Type} & \multirow{2}*{Metrics} & \multicolumn{2}{c}{ConDist} & \multicolumn{2}{c}{CdistR1} & \multicolumn{2}{c}{CdistR2} & \multicolumn{2}{c}{CdistGl} & \multicolumn{2}{c}{All}\\
        &&PLCC&SRCC&PLCC&SRCC&PLCC&SRCC&PLCC&SRCC&PLCC&SRCC\\
        \midrule
        \multirow{4}*{FR-OIQA} 
        & S-PSNR~\cite{yu2015framework} & - & - & 0.2372 & 0.2155 & 0.3588 & 0.2747 & - & - & 0.3023 & 0.2516 \\
        & WS-PSNR~\cite{sun2017weighted} & - & - & 0.2200 & 0.1882 & 0.3546 & 0.2710 & - & - & 0.2952 & 0.2481 \\
        & CPP-PSNR~\cite{zakharchenko2016quality} & - & - & 0.2200 & 0.1882 & 0.3546 & 0.2709 & - & - & 0.2953 & 0.2480 \\
        & WS-SSIM~\cite{zhou2018weighted} & - & - & 0.0764 & 0.0686 & 0.2585 & 0.0942 & - & - & 0.2234 & 0.0622 \\
        \midrule
        \multirow{9}*{2D-IQA}
        & NIQE~\cite{mittal2012making} & 0.1781 & 0.1228 & 0.0753 & 0.0800 & 0.0980 & 0.0671 & 0.4002 & 0.4170 & 0.3677 & 0.2711 \\
        & HyperIQA~\cite{su2020blindly} & 0.2651 & 0.2512 & 0.1641 & 0.1604 & 0.2418 & 0.2392 & 0.5807 & 0.5770 & 0.4750 & 0.4439 \\
        & LinearityIQA~\cite{li2020norm} & 0.3650 & 0.3481 & 0.2288 & 0.2199 & 0.3629 & 0.3539 & 0.6699 & 0.6664 & 0.5659 & 0.5365 \\
        & UNIQUE~\cite{zhang2021uncertainty} & 0.4076 & 0.3988 & 0.1872 & 0.1826 & 0.2422 & 0.2379 & 0.6627 & 0.6668 & 0.5728 & 0.5376 \\
        & CONTRIQUE~\cite{madhusudana2022image} & 0.2101 & 0.1219 & 0.1486 & 0.1423 & 0.2194 & 0.2068 & 0.4174 & 0.4221 & 0.3089 & 0.2625 \\
        & MANIQA~\cite{yang2022maniqa} & 0.0284 & 0.0338 & 0.0038 & 0.0178 & 0.1640 & 0.1552 & 0.2961 & 0.2964 & 0.1129 & 0.0978 \\
        & VCRNet~\cite{pan2022vcrnet} & 0.1866 & 0.1820 & 0.0822 & 0.0696 & 0.2415 & 0.2302 & 0.3409 & 0.3396 & 0.2891 & 0.2772 \\
        & LIQE~\cite{zhang2023blind} & 0.6373 & 0.6505 & 0.0899 & 0.0813 & 0.0750 & 0.0654 & 0.6828 & 0.6226 & 0.5513 & 0.5216 \\
        & SAAN~\cite{yi2023towards} & 0.0622 & 0.0564 & 0.0564 & 0.0575 & 0.1334 & 0.1087 & 0.1301 & 0.0353 & 0.1479 & 0.0359 \\
        \midrule
        \multirow{5}*{NR-OIQA} 
        & MC360IQA~\cite{sun2019mc360iqa} & 0.5018 & 0.4616 & 0.4464 & 0.4237 & 0.6263 & 0.6250 & 0.7823 & 0.7673 & 0.7206 & 0.7099 \\
        & VGCN~\cite{xu2020blind} & 0.4113 & 0.3827 & 0.4979 & 0.4793 & 0.6536 & 0.6491 & 0.7632 & 0.7281 & 0.7057 & 0.6987 \\
        & Fang22~\cite{fang2022perceptual} & 0.5527 & 0.5311 & \underline{0.5424} & \underline{0.5222} & \underline{0.6732} & \underline{0.6695} & 0.8317 & 0.7975 & 0.7692 & 0.7581\\
        & Assessor360~\cite{wu2023assessor360} & \underline{0.6779} & \underline{0.6787} & 0.5084 & 0.4789 & 0.6627 & 0.6517 & \underline{0.8376} & \underline{0.8107} & \underline{0.7904} & \underline{0.7731}\\
        & IQCaption360 & \textbf{0.6876} & \textbf{0.6974} & \textbf{0.6248} & \textbf{0.6134} & \textbf{0.7320} & \textbf{0.7319} & \textbf{0.8502} & \textbf{0.8185} & \textbf{0.8181} & \textbf{0.8147}\\
        \bottomrule
    \end{tabular}
\end{table*}

% \begin{table}[t]
%     \centering
%     \caption{Ablation experiments on OIQ-10K}
%     \label{tab:results2}
%     \begin{tabular}{lccc}
%         \toprule
%         Variants & ACC & PLCC & SRCC\\
%         \midrule
%         Score Distribution & \textbf{0.9550} & 0.8168 & 0.8140 \\
%         Merge Viewports & 0.9315 & 0.8118 & 0.8114 \\
%         Use position embedding (Cat) & \underline{0.9500} & \textbf{0.8200} & \textbf{0.8197} \\
%         Use position embedding (Add) & 0.9445 & 0.8099 & 0.8086 \\
%         IQCaption360 & 0.9480 & \underline{0.8181} & \underline{0.8147} \\
%         \bottomrule
%     \end{tabular}
% \end{table}

\subsection{Experimental Results}
\label{subsec:ex_re}

We compare the performance of the proposed IQCaption360 with the following state-of-the-art models, including four FR-OIQA models: S-SPNR~\cite{yu2015framework}, WS-PSNR~\cite{sun2017weighted}, CPP-PSNR~\cite{zakharchenko2016quality}, WS-SSIM \cite{zhou2018weighted}, nine 2D-IQA models: NIQE~\cite{mittal2012making}, HyperIQA~\cite{su2020blindly}, LiearityIQA~\cite{li2020norm}, UNIQUE~\cite{zhang2021uncertainty}, CONTRIQUE~\cite{madhusudana2022image}, MANIQA~\cite{yang2022maniqa}, VCRNet~\cite{pan2022vcrnet}, LIQE~\cite{zhang2023blind}, SAAN~\cite{yi2023towards}, and four NR-OIQA models: MC360IQA \cite{sun2019mc360iqa}, VGCN~\cite{xu2020blind}, Fang22~\cite{fang2022perceptual} and Assessor360 \cite{wu2023assessor360}. To ensure a fair comparison, we re-train the NR-OIQA metrics on the OIQ-10K database, and 2D-IQA metrics are tested using their pre-trained parameters with the best performance. The comparison results are shown in Table~\ref{tab:sota}, where we have several significant observations. First, all of these FR-OIQA and 2D-IQA metrics exhibit poor performance in the CdistR1 and CdistR2 situations, which indicates the inherent complexity of quality modeling of heterogeneously distorted omnidirectional images. On the contrary, NR-OIQA metrics show better performance owing to their specialized design in modeling the relationship of the viewport sequence. Second, LinearityIQA~\cite{li2020norm}, UNIQUE~\cite{zhang2021uncertainty} and LIQE~\cite{zhang2023blind} present promising results among the 2D-IQA metrics. LinearityIQA is optimized based on the Norm-in-Norm loss, in which the embedded normalization helps to improve the smoothness of the loss landscape for better performance. UNIQUE uses a multi-database training strategy, continuous ranking annotation, and hinge regularizer, while LIQE is based on a multitask learning scheme, including scene classification, distortion type identification, and quality assessment. The multi-database training strategy and multitask learning scheme contribute to improving the model's generalization ability. SAAN~\cite{yi2023towards} shows pool performance on the proposed OIQ-10K database, potentially because SAAN places much emphasis on the aesthetic perspective (\ie, components and color harmony) rather than the technical perspective (\ie, distortion type and degradation degree).

Overall, the proposed IQCaption360 demonstrates better performance against the state-of-the-art metrics. Specifically, IQCaption360 surpasses VGCN~\cite{xu2020blind} by 11.24\% and outperforms Fang22~\cite{fang2022perceptual} by 4.89\% in terms of PLCC. Note that VGCN~\cite{xu2020blind} shows relatively less favorable performance in the CnoDist situation, which may be attributed to the visual sensitivity of humans to regions of inferior quality, particularly when the image is of high quality. Besides, Fang22~\cite{fang2022perceptual} achieves suboptimal results in the CdistR1 and CdistR2 situations (\emph{i.e.}, heterogeneous distortion), which may benefit from the consideration of viewing behavior. Furthermore, IQCaption360 gains 2.77\% improvement in terms of PLCC and 4.16\% in terms of SRCC compared to Assessor360~\cite{wu2023assessor360}. The gap is largely reflected in the CdistR1 and CdistR2 situations, one possible reason is that Assessor360 may not be suitable for handling heterogeneous distortions, as the entropy-based viewport sampling strategy could produce biased sequences due to distortion discontinuities. It should also be noted that IQCaption360 outperforms other competitors by a significant margin in the CdistR1 and CdistR2 situations, which is attributed to that the VPFS module effectively fuses the features of viewports with different perceptual quality.

\begin{figure*}[t]
\centering
\includegraphics[width=0.9\linewidth]{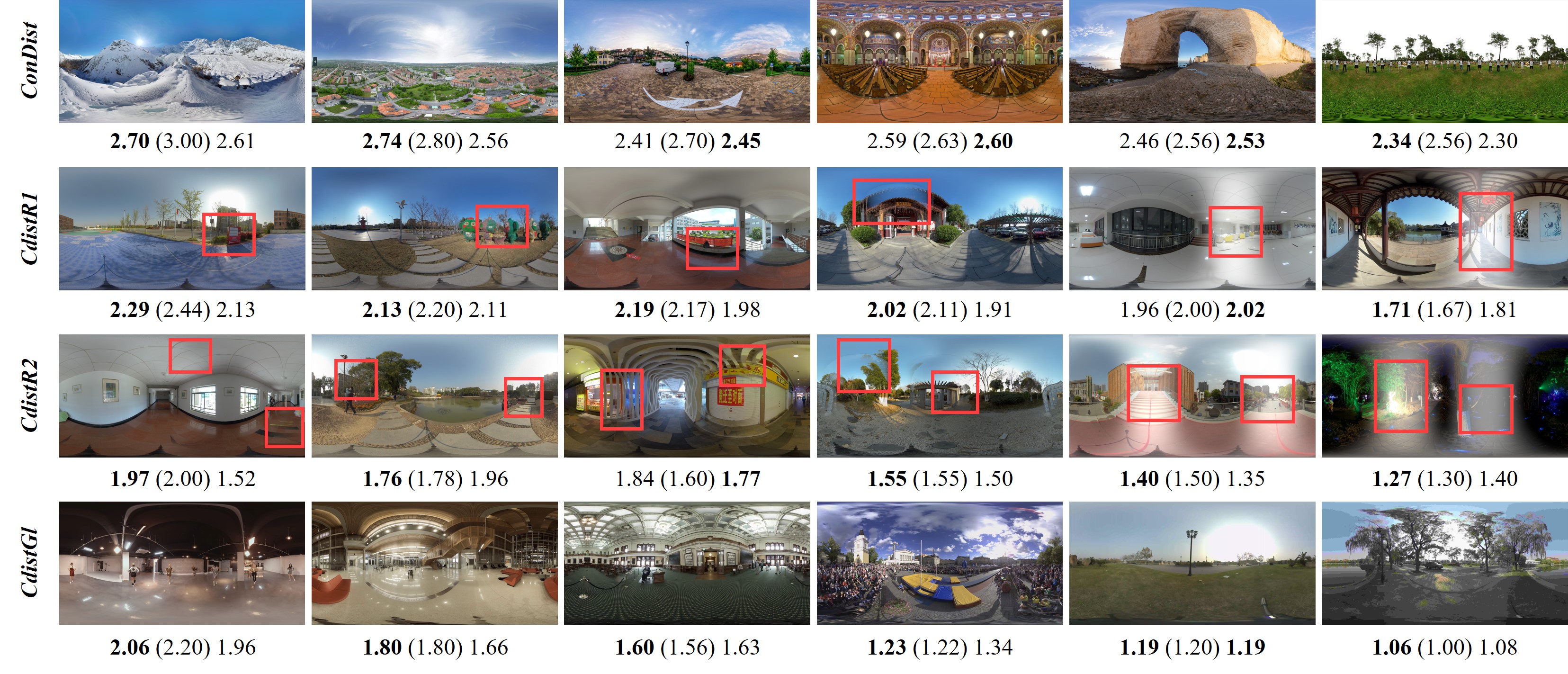}
\caption{Visualization of samples in the OIQ-10K and the corresponding processing results of four distortion situations. The red box marks the obvious distortion area for better presentation. The values from left to right are the prediction of IQCaption360, MOS, and the prediction of Assessor360. The closest value to the MOS is marked in bold.}
\label{fig:vi}
\end{figure*}

\subsection{Visual results}
\label{subsec:vi}
To intuitively show the superiority of the proposed IQCaption360 against the state-of-the-art OIQA model Assessor360, we exhibit some samples and their prediction results across four distortion situations from the OIQ-10K database, as illustrated in Figure~\ref{fig:vi}. We find that IQCaption360 accurately assesses the quality of omnidirectional images in most scenarios, particularly for those with heterogeneous distortion. This demonstrates the effectiveness of the multi-task joint training strategy and the VPFS module. On the other hand, Assessor360 performs well on homogeneously distorted omnidirectional images, however, sometimes it is less accurate when dealing with heterogeneous distortion (\emph{e.g.}, the first image of CdistR1, the first and fifth images of CdistR2). The possible reason is that its entropy-based sampling approach could be affected by distortion discontinuity.

% which may be attributed to the consideration of both auxiliary knowledge and the discrepancy between viewports from the same omnidirectional images

\begin{table}[t]
    \centering
    \caption{Ablation studies on each component. Note that the performance difference of each two of the variants 1, 2, 3, and 4 is statistically significant (Student's t-test with p$<$0.05).}
    \label{tab:modules}
    \begin{tabular}{cccccccc}
        \toprule
        \#&QSPN & DSPN & MSFS & VPFS & ACC & PLCC & SRCC \\
        \midrule
        1&$\checkmark$ &  &  &  & - & 0.7918 & 0.7819 \\
        2&$\checkmark$ & $\checkmark$ &  &  & 0.9340 & \underline{0.8114} & 0.8052 \\
        3&$\checkmark$ & $\checkmark$ & $\checkmark$ &  & \textbf{0.9545} & 0.8111 &\underline{0.8106} \\
        4&$\checkmark$ & $\checkmark$ & $\checkmark$ & $\checkmark$ & \underline{0.9480} & \textbf{0.8181} & \textbf{0.8147} \\
        \bottomrule
    \end{tabular}
\end{table}

\begin{table}[t]
    \centering
    \caption{Ablation study on the VPFS module.}
    \label{tab:select_vp_num}
    \begin{tabular}{lcccc}
        \toprule
        Variants & K & ACC & PLCC & SRCC \\
        \midrule
        All Viewports & 8 & 0.9260 & 0.8109 & 0.8080 \\
        1/2 Viewports & 4 & \underline{0.9480} & \textbf{0.8181} & \textbf{0.8147}\\
        1/4 Viewports & 2 & 0.9450 & \underline{0.8166} & \underline{0.8137} \\
        1/8 Viewports & 1 & \textbf{0.9570} & 0.8123 & 0.8109 \\
        \bottomrule
    \end{tabular}
\end{table}

\begin{table}[t]
    \centering
    \caption{Ablation study on the weights of different tasks. The $Auto$ refers to using the dynamic average method.}
    \label{tab:task_weight}
    \begin{tabular}{ccccc}
        \toprule
        DSPN & QSPN & ACC & PLCC & SRCC \\
        \midrule
        0.1 & 0.9 & 0.9105 & 0.7999 & 0.7955 \\
        0.3 & 0.7 & 0.9450 & 0.8060 & 0.8031 \\
        0.5 & 0.5 & \textbf{0.9495} & \underline{0.8146} & \underline{0.8129} \\
        0.7 & 0.3 & 0.9395 & 0.8144 & 0.8108 \\
        0.9 & 0.1 & 0.9490 & 0.8000 & 0.7979 \\
        Auto & Auto & \underline{0.9480} & \textbf{0.8181} & \textbf{0.8147} \\
        \bottomrule
    \end{tabular}
\end{table}

\begin{table}[t]
    \centering
    \caption{Ablation study on viewport size. $FLOPs$ refers to floating point operations.}
    \label{tab:vp_size}
    \begin{tabular}{crcccc}
        \toprule
        Viewport Size & FLOPs & ACC & PLCC & SRCC \\
        \midrule
        $128\times128$ & 7.9 G & 0.9385 & 0.8022 & 0.7970 \\
        $224\times224$ & 22.2 G & \textbf{0.9480} & \textbf{0.8181} & \textbf{0.8147} \\
        $384\times384$ & 65.2 G & 0.9320 & \underline{0.8107} & \underline{0.8071} \\
        $512\times512$ & 116.0 G & \underline{0.9415} & 0.8024 & 0.7955 \\
        \bottomrule
    \end{tabular}
\end{table}

\begin{table}[t]
    \centering
    \caption{Quantitative comparison of using different viewport generation methods on the OIQ-10K database.}
    \label{tab:vp_gen_met}
    \begin{tabular}{lcccc}
        \toprule
        Generation Methods & Num & ACC & PLCC & SRCC \\
        \midrule
        Spherical Sampling \cite{fang2022perceptual} & 20 & \underline{0.9170} & \underline{0.7850} & \underline{0.7784} \\
        Structural Sampling\cite{xu2020blind} & 20 & 0.8175 & 0.7663 & 0.7604 \\
        RPS \cite{wu2023assessor360} & 15 & 0.8205 & 0.7683 & 0.7546 \\
        ScanDMM \cite{sui2023scandmm} & 15 & 0.7585 & 0.7570 & 0.7387 \\
        Realistic Scanpath & 15 & 0.8110 & 0.4791 & 0.4775 \\
        Equatorial Sampling & 8 & \textbf{0.9480} & \textbf{0.8181} & \textbf{0.8147} \\
        \bottomrule
    \end{tabular}
\end{table}

\subsection{Ablation Studies}
\label{subsec:ab_st}

\textbf{Effectiveness of Each Component.} We verify the effectiveness of each component of the proposed IQCaption360, and the experimental results are shown in Table~\ref{tab:modules}. First, the task of distortion situation identification greatly contributes to OIQA, reflecting the synergistic relationship between them. Moreover, we introduce the MSFS module in the feature aggregation phase of IQCaption360 to adaptively tailor the general features to the task-specific features, yielding performance improvement for both tasks. Additionally, the performance of IQCaption360 is further improved with the implementation of the VPFS module, which indicates that it is of significance to consider the difference between viewports in the modeling omnidirectional image quality.

\textbf{Impact of the Number of Viewports.} To study the impact of the VPFS module, we set various selections of viewport numbers by setting different K, and the results are presented in Table \ref{tab:select_vp_num}. First, we can observe that the results are unsatisfactory if we naively select viewports. Second, selecting half of the viewports can achieve the best performance, while the performance begins to decline as the number of selected viewports continues to decrease. The results imply that more viewports might introduce redundant information, while fewer viewports cause a handle in quality evaluation.

\textbf{Impact of Task Weights.} We also explore the effect of task weights $\lambda_1$ and $\lambda_2$ in adjusting the importance of each task. As shown in Table~\ref{tab:task_weight}, the performance is sub-optimal when assigning the same weights to these two tasks, and other fixed weight settings can also not get promising results. In contrast, we adopt the dynamic weight assignment method~\cite{liu2019end} to automatically allocate weights for each task and obtain the best performance.

\textbf{Impact of Viewport Size.} We further investigate the influence of viewport size on the performance, where we set the viewport size to 128$\times$128, 224$\times$224, 384$\times$384 and 512$\times$512, and the experimental results are presented in Table \ref{tab:vp_size}. Overall, the proposed IQCaption360 behaves stably and obtains the best performance and relatively low FLOPs when the input size is set to 224$\times$224. Moreover, an excess of visual features may introduce redundant information, while an insufficient amount may lead to inadequate information, both of which could potentially disrupt the training process.

\textbf{Impact of Viewport Generation Method.}
Since the projection from ERP to viewports will impair visual information, it is crucial to generate a reasonable set of viewports to represent omnidirectional images in the format of ERP to the greatest extent. Thus, we test the impact of existing viewport generation methods, and the experimental results are shown in Table~\ref{tab:vp_gen_met}. From this table, the simple equatorial sampling method adopted in the IQCaption360 showcases its effectiveness in handling both tasks. It is attributed to the consideration that visual information is concentrated in low-latitude regions and the uniform sampling approach is able to capture global quality~\cite{wu2022fast}. The spherical sampling method~\cite{fang2022perceptual} only takes the global quality into account while the information from the non-equatorial region could introduce redundant features. The structural sampling method~\cite{xu2020blind} generates viewports based on the theory that humans are more sensitive to structural information. However, the structural features in images are undeniably impaired in the degradation process and lead to unreasonable sampling results. Although the sampling approaches~\cite {wu2023assessor360,sui2023scandmm} show dominating performance on the homogeneously distorted OIQ databases, they may overlook the crucial global quality in the generating process, especially for assessing the omnidirectional images with heterogeneous distortion. Note that the scanpath-based approach exhibits inferior performance, the reason could be that our model does not introduce relevant design (\eg, GRUs) for dealing with the temporal dependencies in viewports~\cite{wu2023assessor360,zhang2023blind}.

\subsection{Cross-Database Evaluation}
To test the generalization ability of our IQCaption360, we perform cross-database validation on the MVAQD~\cite{jiang2021cubemap} and IQA-ODI~\cite{yang2021spatial}, and compare it with MC360IQA and Assessor360. The results are listed in Table~\ref{tab:cross}, from which we can observe that IQCaption360 shows superior generalization performance compared to the other metrics. The MVAQD includes five distortion types and four levels, while IQA-ODI is influenced by four projection patterns and distortion levels, presenting challenges for existing OIQA metrics. However, MC360IQA and Assessor360 face challenges in effectively capturing heterogeneous distortion, which limits their ability to fully leverage the distortion knowledge in OIQ-10K to improve generalization ability. In contrast, the proposed IQCaption360 effectively absorbs the rich distortion-related knowledge from the OIQ-10K, bringing better generalization ability.

\begin{table}[!htbp]
    \centering
    \caption{Cross-database evaluation on SRCC results. The OIQA metrics are trained on the proposed OIQ-10K and tested on the MVAQD~\cite{jiang2021cubemap} and IQA-ODI~\cite{yang2021spatial}.}
    \label{tab:cross}
    \begin{tabular}{lcccc}
\toprule
Testing set & MC360IQA & Assessor360 & IQCaption360 \\
\midrule
  MVAQD & 0.5332 &  \underline{0.6801} & \textbf{0.7891} \\
  IQA-ODI & 0.4364 & \underline{0.6281} & \textbf{0.7428} \\
\bottomrule
    \end{tabular}
\end{table}

\subsection{Computational Complexity}
To demonstrate the practical performance of the proposed IQCaption360, we evaluate the computational complexity and speed of our model and other test OIQA models on the OIQ-10K database, the results are shown in Table~\ref{tab:speed}. It is evident that IQCaption360 achieves the fastest inference speed, processing 66 images per second. This is primarily due to applying NAT as the backbone, which operates with a specifically developed Python package that significantly enhances computational speed. Additionally, lightweight designs contribute as well. For instance, most key functions are handled by simple MLP layers except for the backbone.

\begin{table}[!htbp]
    \centering
    \caption{Computational comparison of different models.}
    \label{tab:speed}
    \begin{tabular}{lrrc}
\toprule
Metrics & Params & FLOPs & Speed (FPS) \\
\midrule
  MC360IQA & 22.4 M& 30.3 G& \underline{7}\\
  VGCN & 21.6 M& 191.5 G& 1\\
  Fang22 & 25.2 M& 174.3 G& 2\\
  Assessor360 & 88.2 M& 230.5 G& 4\\
  IQCaption360 & 29.3 M& 22.2 G& \textbf{66} \\
\bottomrule
    \end{tabular}
\end{table}
% zhang2022no,

%-------------------------------------------------------------------------
\section{Conclusion and Future Work}
\label{sec:conc}

In this paper, we construct a new OIQA database called OIQ-10K, which consists of 10,000 omnidirectional images degraded by both homogeneous and heterogeneous distortion. A large-scale psychophysical study is carefully elaborated to collect human opinions of each omnidirectional image to promote the development of OIQA, and we study the influence of distortion situation and viewing condition on perceived quality, where we have several interesting findings: $\mathbf{\romannumeral1)}$ perceptual quality is negatively correlated with the number of distorted regions; $\mathbf{\romannumeral2)}$ subjects have consistent opinions on both high and low-quality omnidirectional images, which is no longer valid for general-quality images; $\mathbf{\romannumeral3)}$ the last impression is more influential than the initial impression for quality assessment. Furthermore, based on the observation that human recognition is mainly based on semantic description rather than solely relying set of related values, we propose a novel multitask-derived adaptive feature-tailoring model named IQCaption360 for producing quality captions of omnidirectional images by combining the predicted results (including distortion situation and quality score) in a textual template. Extensive experiments demonstrate the effectiveness of IQCaption360, and it surpasses the state-of-the-art methods by a significant margin on the proposed OIQ-10K database. 

In the future, we plan to design the end-to-end model to deal with the OIQC problem, rather than relying on the pre-defined templates. A straightforward solution is adapting image captioning~\cite{stefanini2022show} architectures to OIQC with the specific combination of quality-aware features extractor of omnidirectional images, where we can resort to large language models (LLMs) to generate quality-relevant descriptions as the prompts or supervision signals of the OIQC model.

%\appendices
%% you can choose not to have a title for an appendix
%% if you want by leaving the argument blank
%\section{}


% \section*{Acknowledgment}
% The authors would like to thank all participants in the subjective experiment.

\ifCLASSOPTIONcaptionsoff
  \newpage
\fi



%\bibliographystyle{IEEEtran}

\bibliographystyle{IEEEtran}
\bibliography{egbib}

\begin{IEEEbiography}[{\includegraphics[width=1.0in,height=1.3in,clip]{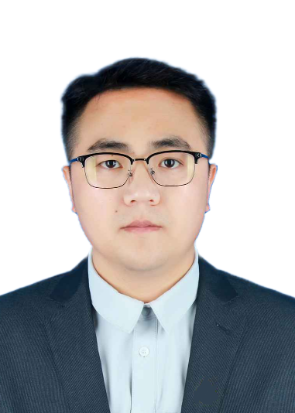}}]{Jiebin Yan} received the Ph.D. degree from Jiangxi University of Finance and Economics, Nanchang, China. He was a computer vision engineer with MTlab, Meitu. Inc, and a research intern with MOKU Laboratory, Alibaba Group. From 2021 to 2022, he was a visiting Ph.D. student with the Department of Electrical and Computer Engineering, University of Waterloo, Canada. He is currently a Lecturer with the School of Computing and Artificial Intelligence, Jiangxi University of Finance and Economics, Nanchang, China. His research interests include visual quality assessment and computer vision.
\end{IEEEbiography}
\vspace{-10 mm} 

\begin{IEEEbiography}[{\includegraphics[width=1.0in,height=1.3in,clip]{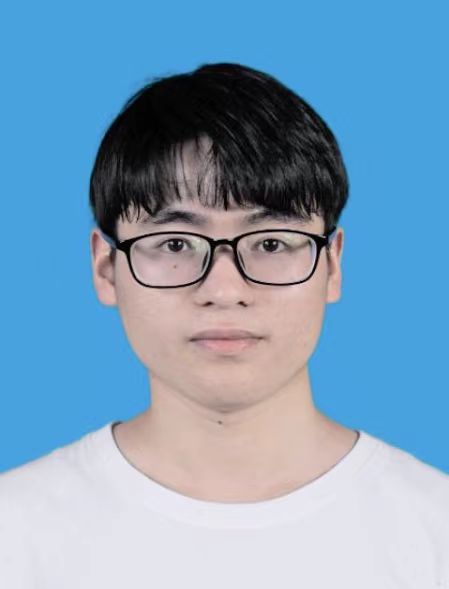}}]{Ziwen Tan} received the B.E. degree from the School of Modern Economics and Management, Jiangxi University of Finance and Economics, Nanchang, China, in 2022. He is currently working toward the M.S. degree with the School of Computing and Artificial Intelligence, Jiangxi University of Finance and Economics. His research interests include visual quality assessment and VR image processing.
\end{IEEEbiography}
\vspace{-10 mm} 

\begin{IEEEbiography}[{\includegraphics[width=1.0in,height=1.2in,clip]{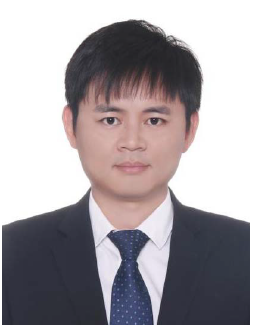}}]{Yuming Fang}(S’13–SM’17) received the B.E. degree from Sichuan University, Chengdu, China, the M.S. degree from the Beijing University of Technology, Beijing, China, and the Ph.D. degree from Nanyang Technological University, Singapore. He is currently a Professor with the School of Computing and Artificial Intelligence, Jiangxi University of Finance and Economics, Nanchang, China. His research interests include visual attention modeling, visual quality assessment, computer vision, and 3D image/video processing. He serves on the editorial board for \textsc{IEEE Transactions on Multimedia} and \textsc{Signal Processing: Image Communication}.
\end{IEEEbiography}
\vspace{-10 mm} 

\begin{IEEEbiography}[{\includegraphics[width=1.0in,height=1.2in,clip]{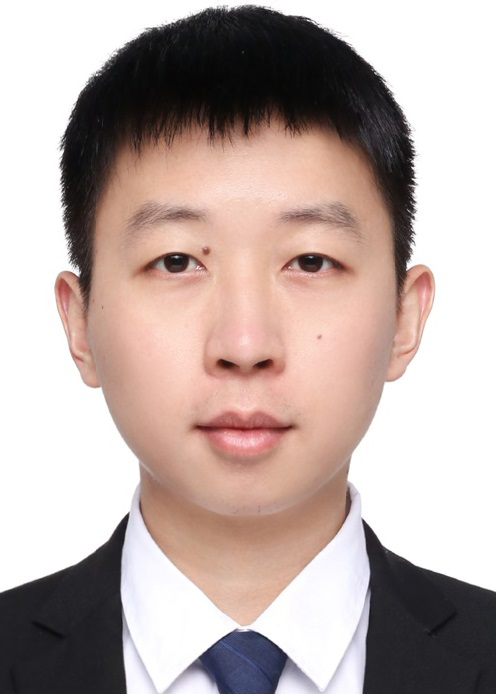}}]{Junjie Chen} received the B.E. degree from Sichuan University in 2018, and received the Ph.D. degree from the Computer Science and Engineering Department, Shanghai Jiao Tong University, Shanghai, China. He is currently a Lecturer with the School of Computing and Artificial Intelligence, Jiangxi University of Finance and Economics, Nanchang, China. His main research interests include computer vision, pose estimation, weakly supervised learning, transfer learning.
\end{IEEEbiography}
\vspace{-10 mm} 

\begin{IEEEbiography}[{\includegraphics[width=1.0in,height=1.2in,clip]{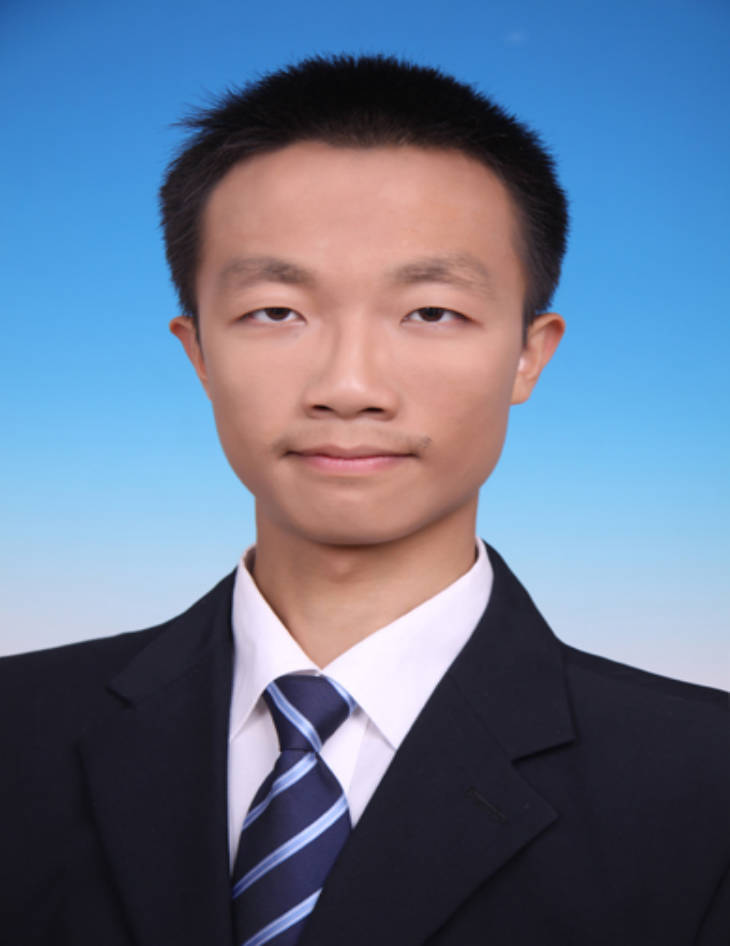}}]{Wenhui Jiang}(Member, IEEE) received the Ph.D. degree from Beijing University of Posts and Telecommunications, Beijing, China. He was a senior engineer with Alibaba Damo Academy from
2017 to 2019, and a visiting student at University of California, Santa Barbara from 2015 to 2016. He is an associate professor with the School of Computing and Artificial Intelligence, Jiangxi University of Finance and Economics, Nanchang, China. His current
research interests include multimedia analysis and deep learning.
\end{IEEEbiography}
\vspace{-10 mm} 

\begin{IEEEbiography}[{\includegraphics[width=1.0in,height=1.2in,clip]{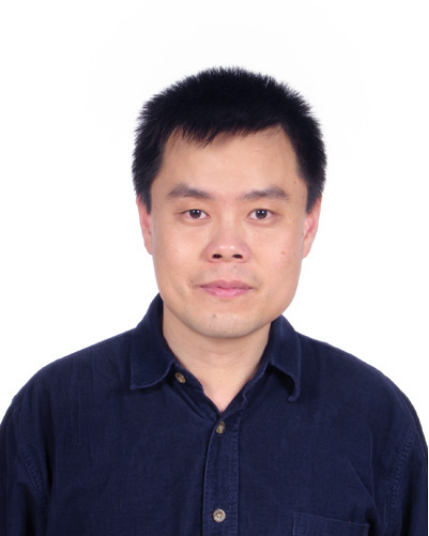}}]{Zhou Wang}(Fellow, EEE) received the Ph.D. degree from the University of Texas at Austin, Austin, TX, USA, in 2001. He is currently a Professor and the Canada Research Chair with the Department of Electrical and Computer Engineering, University of Waterloo, Waterloo, ON, Canada. His research interests include image and video processing and coding visual quality assessment and optimization, computational vision and pattern analysis, multimedia communications, and biomedical signal processing. He has more than 200 publications in these fields with more than 110,000 citations (Google Scholar). Dr. Wang is a Member of the IEEE Multimedia Signal Processing Technical Committee from 2013 to 2015 and IEEE Image, Video and Multidimensional Signal Processing Technical Committee from 2020 to 2022. He was a Fellow of Canadian Academy of Engineering in 2016 and Royal Society of Canada: Academy of Science in 2018. He was the recipient of the 2009 IEEE Signal Processing Society Best Paper Award, 2013 \textit{IEEE Signal Processing Magazine} Best Paper Award, the 2014 NSERC E.W.R. Steacie Memorial Fellowship Award, the 2015 Primetime Engineering Emmy Award, and the 2016 IEEE Signal Processing Society Sustained Impact Paper Award. He serves as a Senior Editor for \textsc{IEEE Journal of Selected Topics in Signal Processing} from 2022 to 2024, a Senior Area Editor of \textsc{IEEE Transactions on Image Processing} from 2015 to 2019, an Associate Editor for \textsc{IEEE Signal Processing Letters} from 2006 to 2010, \textsc{IEEE Transactions on Image Processing} from 2009 to 2014, and \textsc{IEEE Transactions on Circuits and Systems for Video Technology} from 2016 to 2018, and a Guest Editor for \textsc{IEEE Journal of Selected Topics in Signal Processing} from 2013 to 2014 and from 2007 to 2009, among other journals.
\end{IEEEbiography}
\vspace{-10 mm} 

\end{document}